# Title

# Automatic Extraction of Structured Information from Brain MRI Reports Using an Open-Weight Large Language Model

Kaouther Mouheb[1], Amos Pomp[2], Antoine Manenti[1,3], Romy de Haan[4], Farog Faghir[2], Joy Martens[2], Harro Seelaar[4,5], Francesco Mattace- Raso[4,6], Meike W. Vernooij[1,2,4], Frank J. Wolters[1,2,4], Stefan Klein[1], Esther E. Bron[1]

1 Department of Radiology & Nuclear Medicine, Erasmus MC, Rotterdam, the Netherlands
2 Department of Epidemiology, Erasmus MC, Rotterdam, the Netherlands
3 Department of Electrical and Electronics Engineering, ENSEEIHT, Toulouse, France
4 Alzheimer Centre Erasmus MC, Erasmus MC, Rotterdam, the Netherlands
5 Department of Neurology, Erasmus MC, Rotterdam, the Netherlands
6 Department of Internal Medicine, Erasmus MC, Rotterdam, the Netherlands

## Abstract

**Objectives:**
Automatic data extraction from free-text radiology reports enables large-scale research, but few studies assessed the performance of large language models (LLMs) on Dutch neuroradiology reports.
**Methods:**
We analyzed 947 brain MRI reports from a tertiary memory clinic (2016–2021), authored by consultant neuroradiologists. Trained medical students annotated thirty variables; 100 reports were double-annotated to assess inter-rater reliability. We evaluated the performance of the open-weight LLM LLaMA 3.1 using different languages (Dutch vs. English translation) and few-shot prompting with different example selection strategies. Performance was evaluated using balanced accuracy for categorical variables, accuracy and mean absolute error for counts, and text similarity for free-text. Metrics were computed across 10 random splits of the 947 reports.
**Results:**
LLaMA 3.1 demonstrated high zero-shot performance for visual rating scores (mean [95%-CI]): Medial Temporal Atrophy: 90% [77–100%] on the left and 96% [94–99%] on the right, Global Cortical Atrophy: 87% [83–91%], and Fazekas: 94% [93–96%]. Microbleed mentions were detected with 93% accuracy [92–95%] and infarct mentions with 82% [80–84%]. Text similarity for lesion location reached 0.95 [0.95–0.96]. Performance was lower for numerical variables: 80% [78–82%] for the number of microbleeds and 66% [63–68%] for infarcts. English translation yielded comparable results. Few-shot prompting improved performance for numerical variables, achieving 92% [90–93%] for microbleeds and 81% [77–85%] for infarcts using structural similarity-based selection.
**Conclusion:**
LLaMA 3.1 shows strong potential for extracting data from Dutch neuroradiology reports. Few-shot prompting enhances performance for numerical variables, whereas challenges remain for location-specific variables.

## Clinical Relevance Statement

**1. Question:**

The free-text format of radiology reports limits the accessibility of structured data; can large language model LLaMA 3.1 automatically structure free-text Dutch neuroradiology reports?

**2. Findings:**

LLaMA 3.1 accurately extracted 30 variables from Dutch neuroradiology reports, with few-shot prompting significantly improving performance compared to zero-shot prompting, particularly with similarity-based example selection.

**3. Clinical Relevance Statement:**

By enabling accurate, automated post-hoc structuring of free-text radiology reports, open-weight LLMs facilitate large-scale data reuse, improve consistency in reporting, and support clinical research and decision-making in dementia care without disrupting existing clinical workflows.

## Abbreviations

| | |
|---|---|
| CI | Confidence Interval |
| DWI | Diffusion-Weighted Imaging |
| ESR | European Society of Radiology |
| GCA | Global Cortical Atrophy |
| GPT | Generative Pretrained Transformer |
| ICC | Inter-class Correlation |
| IRR | Inter-rater Reliability |
| JSON | JavaScript Object Notation |
| LLM | Large Language Model |
| MAE | Mean Absolute Error |
| MCCV | Monte Carlo Cross-Validation |
| MTA | Medial Temporal Atrophy |
| PACS | Picture Archiving and Communication Systems |
| QC | Quality Control |
| SR | Structured Reporting |
| SWI | Susceptibility-Weighted Imaging |
| WMH | White Matter Hyperintensity |

## Main text

### Introduction

The rapid advancements of radiology techniques and their growing accessibility have led to an increasing volume of radiology reports, offering a rich resource for data-driven medical research. However, their unstructured and free-text format limits their usability, particularly in large-scale analysis [1]. Although expert organizations such as the European Society of Radiology (ESR) advocate for the adoption of structured reporting (SR) in clinical practice, its implementation remains limited [2]. This is primarily due to technical barriers (e.g., incompatibility with radiology information systems), challenges in documenting complex cases, and resistance from radiologists to rigid templates conflicting with their established workflow [3]. Currently, extracting structured data from free-text reports relies on manual annotation which is time-consuming and error prone. This raises the need for automated solutions, particularly for complex applications such as dementia, where reports often contain detailed findings and specialized metrics [4].

Recently, ESR emphasized the potential of large language models (LLMs), such as OpenAI's Generative Pretrained Transformer 4 (GPT-4), in automating the conversion of free-text reports into structured formats [5]. These transformer-based models trained on extensive text corpora can perform advanced language tasks with minimal annotated data [6, 7]. Recent studies demonstrated their effectiveness in extracting clinical information across diverse languages and applications [4, 8]. For instance, Lehnen et al. [9] utilized GPT-3.5 and GPT-4 to extract structured data from stroke reports. Adams et al. [10] evaluated GPT-4 for standardizing German MRI/CT reports, and Sasaki et al. [11] applied GPT-4 to Japanese interventional radiology reports. Cozzi et al. [12] evaluated the performance of GPT-3.5, GPT-4 and Bard in performing BI-RADS classification in Italian, English and Dutch reports. Despite their promising results, significant challenges remain in implementing LLMs for SR [4]. Most studies relied on proprietary models like GPT-4, which require data transfer to external servers, raising privacy concerns and limiting clinical adoption [8]. Open-weight LLMs, which allow local deployment, present a potential solution, though their performance in specialized medical domains requires further validation [13]. In this regard, Nowak et al. [1] reported that open-weight models achieve competitive performance to private models in extracting variables from English chest X-ray reports. Builtjes et al. showed that open-weights models outperform fine-tuned task-specific models on 28 extraction tasks from Dutch medical reports [14]. However, benchmark analyses reveal that such models underperform proprietary counterparts [15], particularly in non-English languages and specialized domains, as their training predominantly relies on general-domain English data [16, 17]. While fine-tuning with domain-specific data can enhance accuracy, hospital settings often lack the required computational resources and annotated datasets [13]. Few-shot prompting, where models are given input-output pairs as examples, emerges as a low-resource alternative that showed promise in simpler tasks such as LI-RADS classification [18] yet its potential in radiology remains underexplored [19, 20].

This study presents a comprehensive evaluation of the open-weight LLM LLaMA 3.1 for structuring Dutch neuroradiology reports, encompassing 30 variables of diverse types (categorical, numerical, and free-text). We first compare its zero-shot performance to human raters using consensus annotations and assess inter-rater reliability. Next, we assess whether English translation of the input reports improves performance. We also investigate few-shot prompting as a resource-efficient alternative to fine-tuning. To our knowledge, this represents the first systematic assessment of an open-weight LLM for structured radiology reporting in Dutch using few-shot prompting, providing critical insights into the feasibility of scalable, privacy-preserving solutions for clinical deployment.

# Materials and Methods

## Dataset

The study was approved by the institutional Medical Ethics Committee with a waiver of informed consent (METC-2023-0569). It retrospectively included Dutch neuroradiology reports from 947 patients who underwent brain MRI as part of their diagnostic work-up at a tertiary memory clinic between 2016 and 2021. We included all consecutive patients with an MRI performed within three months of their initial consultation. Reports were authored as part of routine practice by different radiologists at the same clinic. Of all 947 reports, 152 were drafted following a structured template with predefined fields; the other 795 reports had no predefined structure, leaving the outline of the report up to the radiologist. Each report was manually annotated for 30 variables by one of three trained final-stage medical students, using a structured template with standardized guidelines that defined variable definitions, acceptable values, and provided instructions for handling missing data (See Appendix B for variable definitions and distributions). The template included 24 clinical variables related to brain atrophy and vascular lesions, and 6 internal quality control (QC) variables. We applied Monte Carlo cross-validation (MCCV) by generating 10 random splits of the dataset, each with 50% training and 50% test data. Training sets were used to select few-shot examples; test sets were used for evaluation. To assess inter-rater agreement and compare the model to human performance, a random subset of 100 reports was independently annotated by two raters, with consensus reached in consultation with a neuroradiologist (Author9[1], 15 years experience).

## Large language model prompting

As LLM, we used LLaMA 3.1 (Meta, July 2024) [21], because: (1) its open-weight nature allows local deployment, which complies with hospital data governance requirements; (2) an evaluation on a general-domain Dutch benchmark showed it can process Dutch text, with no performance gain by fine-tuning [22]; and (3) the availability of multiple model sizes (8B, 70B, 405B parameters) allows institutions to choose models that fit their resources. We used the 8B and 70B versions. The 405B version was excluded due to computational constraints.

The prompt was adapted from Lehnen et al. [9] and followed Meta's prompt engineering guidelines[2] (Fig. 1). It included extraction guidelines, a JSON template defining each variable, and the free-text report. The prompt was optimized on 30 reports using 7 variables before running the full inference. The final prompt is provided in Appendix A. Postprocessing consisted of isolating the model-generated text from the output via string parsing and replacing responses that failed JSON validation with 'missing'. Non-conforming categorical outputs (e.g., Fazekas score of 5) were assigned an *invalid* label. The analysis was implemented in Python 3.11[3] using HuggingFace's Transformers library (v4.48.3)[4] and executed on two NVIDIA H100 GPUs. To ensure reproducibility, we disabled sampling (i.e., used greedy decoding) to eliminate stochasticity. The code is publicly available: https://anonymous.4open.science/r/LLaMA4Radiology-B006. The dataset can be requested by contacting the authors.

[1] Author's initials will be added after blinded review
[2] https://www.LLaMA.com/docs/how-to-guides/prompting/
[3] https://www.python.org/downloads/release/python-3110/
[4] https://huggingface.co/docs/transformers/en/index

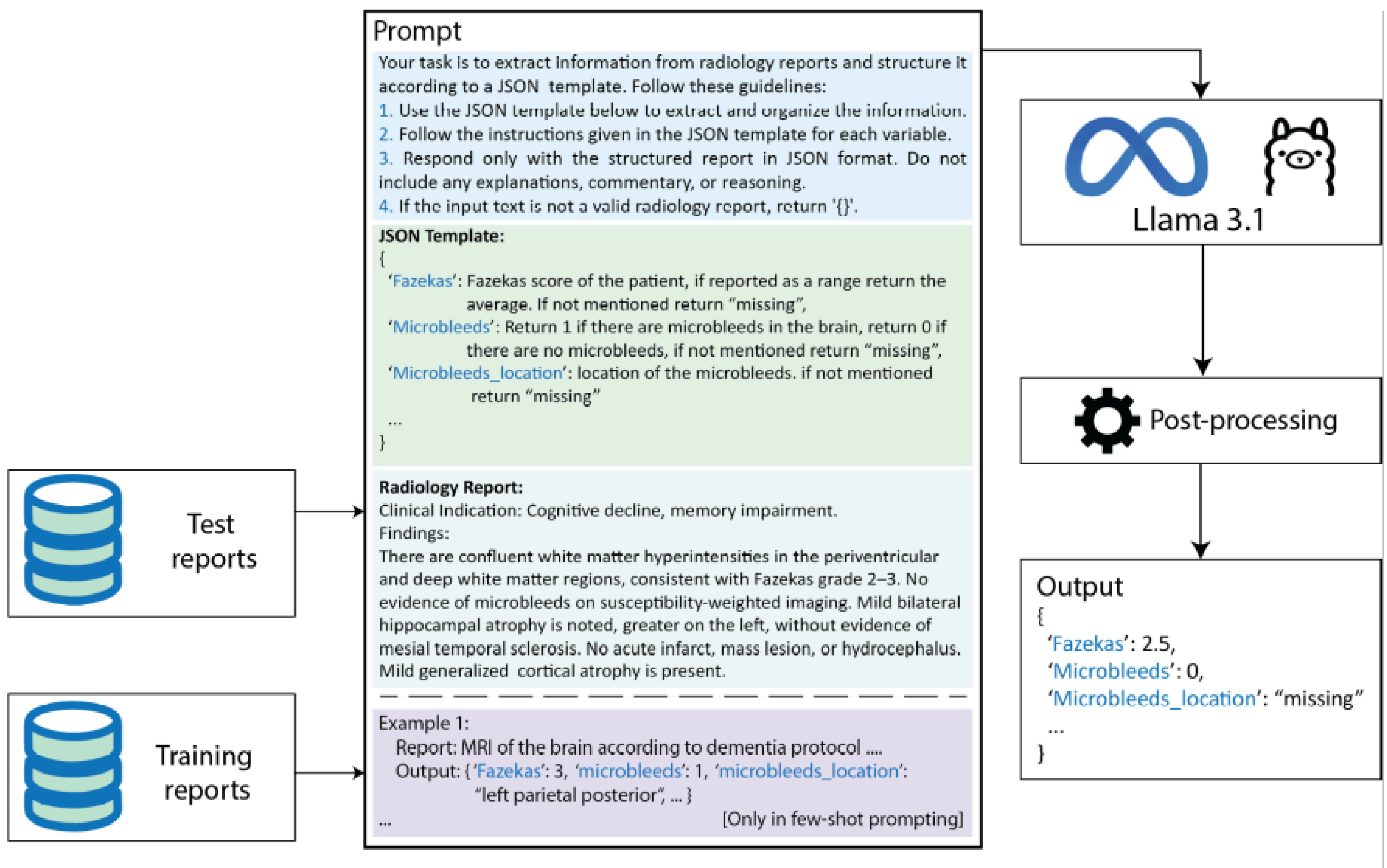

Figure 1: Graphical illustration of the data extraction pipeline using LLaMA 3.1.

## Experimental design

We perform the following analyses:

**1. Comparison to human performance and inter-rater agreement:** Using the dual-annotated Dutch reports, we evaluated the model's zero-shot performance (i.e. its ability to extract information without task-specific examples), compared to human performance. Model outputs and individual human annotations were compared to consensus labels as reference standard. In parallel, we assessed inter-rater reliability (IRR) between the two annotators.

**2. Impact of input language:** We evaluated whether translating Dutch radiology reports to English impacts the model's zero-shot performance. Three conditions were compared: original Dutch reports, English translations generated by LLaMA 3.1, and translations generated by a Helsinki NLP's OPUS-MT model [23]. For English inputs, extracted text fields were back-translated into Dutch using LLaMA 3.1 to maintain evaluation consistency. An author (Author1[5]) rated translations of 50 reports as good, acceptable, or poor, and repeated the ratings 8 months later to assess consistency.

**3. Few-shot prompting:** To evaluate the impact of few-shot prompting, three example reports from the training set with their manually structured counterparts were given in the input prompt to guide output generation. Experiments were conducted using the original Dutch reports. Four example selection strategies were compared:

[5] Author's initials will be added after blinded review

1. *Random Selection*: different examples are randomly sampled from the training set for each report.
2. *Fixed Selection*: A manually curated set is selected from the training set of each split ensuring coverage of commonly occurring values, rare cases, and reports with missing information.
3. *Structural Similarity*: For each candidate training report, we computed its Jaccard similarity with the test report as:

$$J\left(R_{train}, R_{test}\right) = \frac{|S\left(R_{train}\right) \cap S(R_{test})|}{|S\left(R_{train}\right) \cup S(R_{test})|}$$

   Where $R$ represents a report and $S$ is the set of unique words present in it. The three training reports with the highest similarity to the test sample were selected as examples. This measure identifies reports with overlapping terminology and structure.
4. *Semantic Similarity*: For each train report, a semantic text similarity metric with the test report is computed based on the cosine similarity between text embeddings generated using LLaMA 3.1:

$$Cos\left(R_{train}, R_{test}\right) = \frac{f\left(T_{train}\right) \cdot f(T_{test})}{\|f\left(T_{train}\right)\|_2 \|f\left(T_{test}\right)\|_2}$$

   Where R is a report and $T$ is its tokenized vector and $f$ is the encoder of LLaMA 3.1. The three training reports with the highest similarity were selected as examples. This measure identifies reports with similar semantics rather than structure.

## Evaluation metrics and statistical analysis

Variable-specific performance metrics were used. For categorical variables, we used balanced accuracy to account for class imbalance; 'missing' was treated as a separate class. For count variables, we used accuracy (i.e., percentage of model outputs matching the reference annotation) and mean absolute error (MAE). For MAE, entries marked as missing by either the model or the annotator were excluded. For free-text fields, semantic similarity was computed between model outputs and reference annotations by calculating the cosine similarity of sentence embeddings generated using Sentence-BERT [24], with RobBERT-v2 (a Dutch transformer model) as a backbone [25]. Embedding-based metrics are well-established in literature and align closely with human judgment [26]. McNemar's test (α=0.05) [27] was used to evaluate whether the model and human annotators exhibit similar performance to the consensus. Bonferroni correction was applied to control for multiple comparisons. IRR was measured using Cohen's Kappa for categorical variables, intraclass correlation (ICC) for count variables, and semantic similarity for text fields [28]. Agreement levels for Kappa and ICC were defined according to McHugh [29] and Koo and Li [30], respectively. For text similarity, values above 0.9 were considered to indicate high agreement. A bootstrapping and sensitivity analysis is presented in Appendix E to assess the impact of inter-rater differences on the computed metrics. In the remaining experiments, 95% confidence intervals (CIs) were computed following the corrected resampled t-test [31] to account for dependency between the MCCV splits. Statistical differences between prompting methods were evaluated using non-parametric tests. For each variable-method combination, performance metrics were averaged across the 10 splits and converted into ranks, with 1 indicating the best performance. The Friedman test was applied to assess overall differences across methods [32]. When significant, pairwise comparisons were performed using the Nemenyi post-hoc test to identify which methods differed [33].

**Results**

We evaluated the performance of LLaMA 3.1 in extracting 24 neuroradiology report variables. For results related to QC variables, see Appendix C. This section reports results for the 70B version, which performs comparably or significantly better than the 8B version for nearly all variables (Appendix D). F1 scores for categorical variables and Spearman's correlation for numerical fields are provided as additional metrics in Appendix G. Tables in this section are visualized as radar plots in Appendix J.

## Comparison to human performance and inter-rater reliability

Table 1 shows the results of the comparison with human performance and IRR on the 100 double-annotated reports.

LLaMA 3.1 demonstrated high performance on standardized visual rating scales, achieving balanced accuracies of 98% for left MTA, 100% for right MTA, 100% for Fazekas score, and 85% for overall GCA, all comparable to human-level performance ($p$>0.05). Detection of the mention of brain infarcts (83%) and microbleeds (86%) was slightly lower but remained on par with at least one annotator. The model excelled in extracting lesion locations, with high text similarities (0.95 for infarcts, 0.96 for microbleeds). Numerical variables proved challenging, particularly infarct counts, where extraction accuracy ranged from 66% to 78%, falling significantly below Annotator 1 ($p$<0.05). The low MAE suggests that when numerical values were extracted, they were mostly correct; and the lower accuracy is due to confusion between true lesion counts and cases where the count was missing from the report. Performance was also lower for lobar GCA scores and infarct subtypes. In 3 of the 24 variables (occipital GCA, cerebellar infarcts, and splinter infarcts) the model performed significantly worse than both human annotators ($p$<0.05). Notably, these variables were reported less frequently (see Table S1). IRR varied, ranging from near-perfect (κ=0.99 for MTA) to poor agreement (κ=0.39 for superficial siderosis). Considering variables with moderate or higher agreement as reliable, most categorical (15/17) and numerical (4/5) variables and both text fields met this threshold, indicating satisfactory annotation consistency. Bootstrapping results in Appendix E.1 further show that single-rater annotations yield comparable outcomes to consensus annotations. Stratified analysis across agreement levels (Appendix E.2) showed no significant association between IRR and balanced accuracy for categorical variables (Spearman ρ = 0.26, 95% CI [−0.34, 0.77], p = 0.31), indicating that model performance is not systematically lower on variables with weaker inter-rater agreement.

Table 1: Quantitative performance of the LLaMA 3.1 against two human raters and the inter-rater reliability of the two raters on each variable.

| | **LLaMA 3.1** | **Annotator 1** | **Annotator 2** | **Inter-rater reliability** |
|---|---|---|---|---|
| **Categorical Variables** | **Balanced Accuracy (%)** | | | **Cohen's κ** |
| **Atrophy Measures** | | | | |
| MTA left | 100 | 99 | 100 | 0.99 |
| MTA right | 98 | 98 | 100 | 0.99 |
| GCA overall | 85 | 94 | 100 | 0.95 |
| GCA frontal | 74 | 97 | 95 | 0.94 |
| GCA temporal | 72 | 98 | 96* | 0.92 |
| GCA occipital | 61 | 98* | 94* | 0.94 |
| GCA parietal | 70 | 98 | 96 | 0.92 |
| **Vascular Markers** | | | | |
| Fazekas score | 100 | 100 | 100 | 1.0 |

| Any brain infarct | 83 | 90 | 91 | 0.81 |
|---|---|---|---|---|
| Cortical infarcts | 81 | 91* | 81 | 0.59 |
| Lacunes | 88 | 97* | 95 | 0.85 |
| Cerebellar infarcts | 75 | 86* | 91* | 0.68 |
| Splinter infarcts | 68 | 83* | 95* | 0.73 |
| Microbleeds | 86 | 90 | 96* | 0.73 |
| SWI abnormalities | 85 | 95 | 75 | 0.65 |
| Superficial siderosis | 100 | 98 | 100 | 0.39 |
| DWI abnormalities | 91 | 76 | 92 | 0.79 |
| **Numerical Variables** | **Accuracy (%) [Mean Absolute Error]** | | | **ICC** |
| Any brain infarct | 66 [0.08] | 87 [0.06]* | 86 [0.1] | 0.72 |
| Cortical infarcts | 74 [0.0] | 98 [0.06]* | 79 [0.0] | 0.65 |
| Lacunes | 78 [0.12] | 97 [0.05]* | 89 [0.0] | 0.45 |
| Cerebellar infarcts | 73 [0.0] | 90 [0.06]* | 91 [0.0] | 0.83 |
| Microbleeds | 85 [0.0] | 89 [0.03] | 92 [0.0] | 0.97 |
| **Location Variables** | **Text Similarity** | | | **Text Similarity** |
| Infarcts | 0.95 | 1.0 | 0.96 | 0.96 |
| Microbleeds | 0.96 | 0.99 | 0.94 | 0.93 |

ICC: Intra-class correlation; (*) The rater significantly outperforms LLaMA 3.1 based on the McNemar test ($p < 0.05$). A consensus rating was used as a reference standard.

## Impact of input language

Table 2 compares model extraction performance using Dutch reports to their English translations evaluated on the 10 splits of the full dataset.

For Dutch reports, the model showed high balanced accuracies in extracting atrophy measures: 90% (95%-CI: 77-100%) for left MTA, 96% (95%-CI: 94-99%) for right MTA, and 87% (95%-CI: 83-91%) for overall GCA. For vascular markers, the model achieved 94% (95%-CI: 77-100%) for Fazekas score and 93% (95%-CI: 92-95%) for microbleed mention. Infarct mention detection was less accurate (82%, 95%-CI: 80-84%). Lesion counts were more challenging, with an extraction accuracy of 80% (95%-CI: 78-82%) for microbleeds and 66% (95%-CI: 63-68%) for infarcts. Location extraction performance was high for both infarcts and microbleeds (similarity=0.95, 95%-CI: 0.95–0.96). LLaMA 3.1's English translations achieved comparable performance for most variables except for splinter infarcts (translation: 37%, 95%-CI: 31–44 vs. original: 66%, 95%-CI: 64–67%). The term 'splinter infarcts' is a literal translation of the Dutch '*splinterinfarcten*', a descriptive term used locally for small, linear or streak-like, often cortical, cerebellar infarcts. During translation, the model frequently rendered this term as "*small infarct*," thereby losing the morphological specificity required for correct classification. Prompting the model to translate the term as "splinter infarct" improved performance to 65% (95%-CI: 59-71%) without impacting other variables (Appendix F). Across 13 of the 26 variables, using the original Dutch reports outperformed OPUS-MT translations. We observed a similar trend when using F1-score and Spearman's correlation as metrics (Table S9). Translation quality assessment confirmed the superiority of LLaMA 3.1, with 94% of its outputs rated as good, compared to 26% for OPUS-MT (test-retest Cohen's κ=0.7). OPUS-MT often generated incomplete translations and mistranslated medical terms. Both translation methods yielded lower similarity for text fields than those extracted from the Dutch reports.

Table 2: Zero-shot information extraction performance on original Dutch reports vs. English translations by LLaMA 3.1 and OPUS-MT.

| | **Dutch report** | **LLaMA 3.1 Translation** | **OPUS-MT Translation** |
|---|---|---|---|
| **Categorical Variables (%)** | **Balanced Accuracy (%)** | | |
| **Atrophy Measures** | | | |
| MTA left | 90 (77, 100) | 90 (77, 100) | 74 (65, 84) |
| MTA right | 96 (94, 99) | 96 (94, 99) | 79 (70, 88) |
| GCA overall | 87 (83, 91) | 88 (84, 91) | 79 (75, 82) |
| GCA frontal | 80 (73, 87) | 81 (75, 87) | 70 (64, 76) |
| GCA temporal | 73 (69, 78) | 78 (73, 82) | 65 (61, 69) |
| GCA occipital | 58 (47, 69) | 61 (51, 72) | 53 (46, 60) |
| GCA parietal | 77 (70, 84) | 78 (73, 83) | 65 (61, 70) |
| **Vascular Markers** | | | |
| Fazekas score | 94 (93, 96) | 94 (93, 96) | 78 (73, 82) |
| Any brain infarct | 82 (80, 84) | 81 (79, 84) | 66 (63, 70) |
| Cortical infarcts | 75 (69, 81) | 74 (67, 81) | 62 (56, 68) |
| Lacunes | 83 (80, 86) | 85 (82, 88) | 64 (60, 68) |
| Cerebellar infarcts | 69 (66, 71) | 68 (65, 71) | 61 (56, 67) |
| Splinter infarcts | 66 (64, 67) | 37 (31, 44) | 55 (43, 68) |
| Microbleeds | 93 (92, 95) | 94 (92, 95) | 64 (60, 68) |
| SWI abnormalities | 70 (61, 79) | 75 (68, 82) | 54 (47, 62) |
| Presence of siderosis | 97 (85, 100) | 94 (80, 100) | 71 (52, 91) |
| DWI abnormalities | 89 (81, 96) | 85 (77, 94) | 66 (57, 75) |
| **Numerical Variables** | **Accuracy (%) [Mean Absolute Error]** | | |
| Any brain infarct | 66 (63, 68) [0.04] | 67 (63, 70) [0.04] | 59 (56, 62) [0.13] |
| Cortical infarcts | 61 (58, 64) [0.01] | 62 (59, 66) [0.01] | 56 (52, 59) [0.04] |
| Lacunes | 69 (66, 72) [0.10] | 75 (71, 78) [0.08] | 54 (51, 58) [0.09] |
| Cerebellar infarcts | 61 (58, 64) [0.03] | 62 (58, 67) [0.03] | 57 (52, 62) [0.09] |
| Microbleeds | 80 (78, 82) [0.00] | 81 (78, 84) [0.00] | 57 (53, 62) [0.02] |
| **Location Variables** | **Text Similarity** | | |
| Infarcts | 0.95 (0.95, 0.96) | 0.93 (0.92, 0.93) | 0.92 (0.92, 0.93) |
| Microbleeds | 0.95 (0.95, 0.96) | 0.91 (0.90, 0.91) | 0.91 (0.90, 0.91) |

Scores are reported as: value (95%-CI). Underlined means using the translated reports provides significantly worse performance than using the Dutch reports.

## Few-shot prompting

Table 3 presents the results using few-shot prompting evaluated on the 10 splits of the full dataset.

The Friedman test indicated significant differences among prompting methods ($p<0.001$). Nemenyi post-hoc analyses found that all few-shot prompting methods significantly outperformed the zero-shot setting ($p<0.01$) except random few-shot selection ($p=0.85$). Additionally, structural similarity-based selection significantly outperformed random selection ($p<0.001$). No other significant differences were observed for the remaining pairs of methods. In terms of average rank, structural similarity-based selection achieved the lowest rank (1.5) indicating the best overall performance whereas zero-shot prompting ranked last (4.29). For atrophy measures, few-shot prompting slightly increased performance for GCA scores. For vascular markers, a substantial improvement was observed in the detection of lesion presence, particularly using structural similarity-based selection (balanced accuracy for cerebellar infarcts: 85%, 95%-CI: 81-89%; splinter infarcts: 86%, 95%-CI: 82-89%). Random and fixed selection increased the extraction accuracy for three vascular lesion counts (cortical infarcts, lacunes, cerebellar infarcts) and similarity-based approaches yielded significant improvements across all numerical variables. For instance, structural similarity-based prompting yielded an accuracy of 81% (95%-CI: 77-85%) for infarct count, exceeding zero-shot performance (66%, 95%-CI: 63-68%). Accuracy for microbleed counts using this method (93% , 95%-CI: 90-96%) was also higher than the zero-shot setting (80%, 95%-CI: 78-82%). Free-text variables showed no improvement.

We compared the examples extracted by the structural and semantic similarity measures (Fig. S9) and found that they extracted different examples for most test reports. Table 4 shows that all methods require under 30s for inference. Few-shot prompting increases inference time by 7%. Random selection is the fastest method at 0.3ms, while semantic similarity-based selection is the slowest at 170ms.

In terms of format, all generated outputs across all settings were syntactically valid JSON strings. The model was instructed to return an empty JSON object if the input did not correspond to a valid radiology report. We quantified the number of empty JSON outputs per experiment. The zero-shot setting produced 13.0 empty JSON outputs per split on average. Few-shot prompting reduced this number substantially: random selection (9.8), fixed selection (5.1), structural similarity-based selection (1.9), and semantic similarity-based selection (2.0). In all non-empty JSON outputs, every key defined in the template (Appendix A) was consistently included in the generated JSON object.

Table 3: Few-shot information extraction performance using fixed examples, randomly selected examples, structural (Jaccard) similarity-based examples and semantic similarity-based examples.

| **Selection method** | **Zero-shot** | **Random** | **Fixed** | **Structural Similarity** | **Semantic Similarity** |
|---|---|---|---|---|---|
| **Categorical Variables** | **Balanced Accuracy (%)** | | | | |
| **Atrophy Measures** | | | | | |
| MTA left | 90 (77, 100) | 91 (77, 100) | 92 (78, 100) | 92 (79, 100) | 91 (79, 100) |
| MTA right | 96 (94, 99) | 95 (89, 100) | 96 (82, 100) | 98 (95, 100) | 96 (89, 100) |
| GCA overall | 87 (83, 91) | 84 (79, 90) | 90 (84, 96) | 89 (85, 93) | 88 (83, 92) |
| GCA frontal | 80 (73, 87) | 78 (72, 84) | 86 (74, 98) | 87 (81, 92) | 84 (75, 93) |
| GCA temporal | 73 (69, 78) | 73 (65, 80) | 81 (66, 96) | 82 (77, 87) | 80 (73, 87) |
| GCA occipital | 58 (47, 69) | 60 (50, 71) | 68 (47, 89) | 69 (59, 80) | 69 (56, 82) |
| GCA parietal | 77 (70, 84) | 76 (71, 82) | 85 (72, 97) | 86 (80, 92) | 83 (75, 91) |
| **Vascular Markers** | | | | | |
| Fazekas score | 94 (93, 96) | 95 (93, 97) | 95 (93, 97) | 95 (92, 97) | 95 (92, 97) |
| Any brain infarct | 82 (80, 84) | 82 (78, 87) | 85 (77, 92) | 88 (83, 93) | 86 (82, 90) |
| Cortical infarcts | 75 (69, 81) | 78 (71, 85) | 81 (75, 87) | 83 (78, 89) | 82 (77, 88) |
| Lacunes | 83 (80, 86) | 85 (80, 89) | 85 (78, 92) | 89 (86, 93) | 88 (85, 91) |
| Cerebellar infarcts | 69 (66, 71) | 75 (70, 79) | **79 (74, 84)** | **85 (81, 89)** | **80 (76, 85)** |
| Splinter infarcts | 66 (64, 67) | **75 (70, 79)** | **80 (74, 86)** | **86 (82, 89)** | **82 (77, 86)** |
| Microbleeds | 93 (92, 95) | 95 (94, 96) | 95 (93, 97) | 95 (93, 96) | 94 (93, 96) |
| SWI abnormalities | 70 (61, 79) | 70 (61, 78) | 71 (46, 97) | 72 (63, 80) | 69 (57, 81) |
| Superficial Siderosis | 97 (85, 100) | 96 (84, 100) | 96 (84, 100) | 97 (85, 100) | 97 (85, 100) |
| DWI abnormalities | 89 (81, 96) | 89 (81, 97) | 89 (81, 96) | 91 (83, 98) | 91 (83, 98) |
| **Numerical Variables** | **Accuracy (%) [Mean Absolute Error]** | | | | |
| Any brain infarct | 66 (63, 68) [0.04] | 73 (68, 78) [0.05] | 76 (68, 85) [0.04] | **81 (77, 85)** [0.05] | **78 (74, 83)** [0.06] |
| Cortical infarcts | 61 (58, 64) [0.01] | **72 (67, 77)** [0.02] | **77 (70, 84)** [0.02] | **81 (78, 85)** [0.02] | **80 (75, 85)** [0.02] |
| Lacunes | 69 (66, 72) [0.10] | **80 (75, 84)** [0.07] | **81 (73, 89)** [0.07] | **86 (82, 90)** [0.07] | **84 (80, 88)** [0.07] |
| Cerebellar infarcts | 61 (58, 64) [0.03] | 65 (59, 70) [0.03] | 69 (63, 76) [0.03] | **78 (72, 84)** [0.03] | **73 (67, 78)** [0.04] |
| Microbleeds | 80 (78, 82) [0.00] | **92 (90, 95)** [0.01] | **93 (90, 96)** [0.01] | **92 (90, 93)** [0.01] | **92 (90, 94)** [0.01] |
| **Location Variables** | **Text Similarity** | | | | |
| Infarcts | 0.95 (0.95, 0.96) | 0.95 (0.95, 0.96) | 0.95 (0.95, 0.95) | 0.95 (0.95, 0.96) | 0.95 (0.95, 0.96) |
| Microbleeds | 0.95 (0.95, 0.96) | 0.95 (0.95, 0.95) | 0.95 (0.95, 0.96) | 0.95 (0.95, 0.95) | 0.95 (0.94, 0.95) |

Scores are reported as: value (95%-CI). Entries where performance is significantly higher than the zero-shot performance are highlighted in **bold**.

## Analysis of common mistakes

To analyze errors, Fig. 2 presents confusion matrices obtained using structural similarity-based few-shot prompting for four categorical variables (left MTA, overall GCA, brain infarcts,microbleeds) and two counts (numbers of microbleeds and brain infarcts); remaining confusion matrices are in Appendix I. Appendix H shows qualitative examples.

The confusion matrices showed that most misclassifications involve confusion between the 'missing' class and explicit negative labels (e.g., "No," "Absent") for categorical variables and between 'missing' and zero for counts. This indicates that the model frequently fails to differentiate cases where an abnormality is explicitly ruled out from those in which the relevant information is not provided. A sensitivity analysis (Appendix L) confirmed that this error mode has an impact on performance where merging 'missing' into the negative class increased balanced accuracy by up to 14 percentage points for variables with a high prevalence of missing labels (e.g., cerebellar and splinter infarcts), while performance was robust (within 3%) for variables with fewer missing labels. For ordinal scoring variables, errors predominantly occur when reports express uncertainty between adjacent scores (e.g., a GCA score of “1 to 2” is most likely to be misclassified as either “1” or “2”). The model occasionally exhibits hallucinations, such as assigning scores to entries labeled as “missing” or predicting values outside the permissible range, this was mainly due to confusion between concepts (see Microbleeds information in example 3; Table S11), or the model inferring scores based on its previous knowledge even when the radiologist does not specify a score (see MTA in example 4; Table S11). Qualitative analysis further identified errors attributable to inconsistencies in manual annotation, particularly for variables with low IRR (e.g., infarct presence).

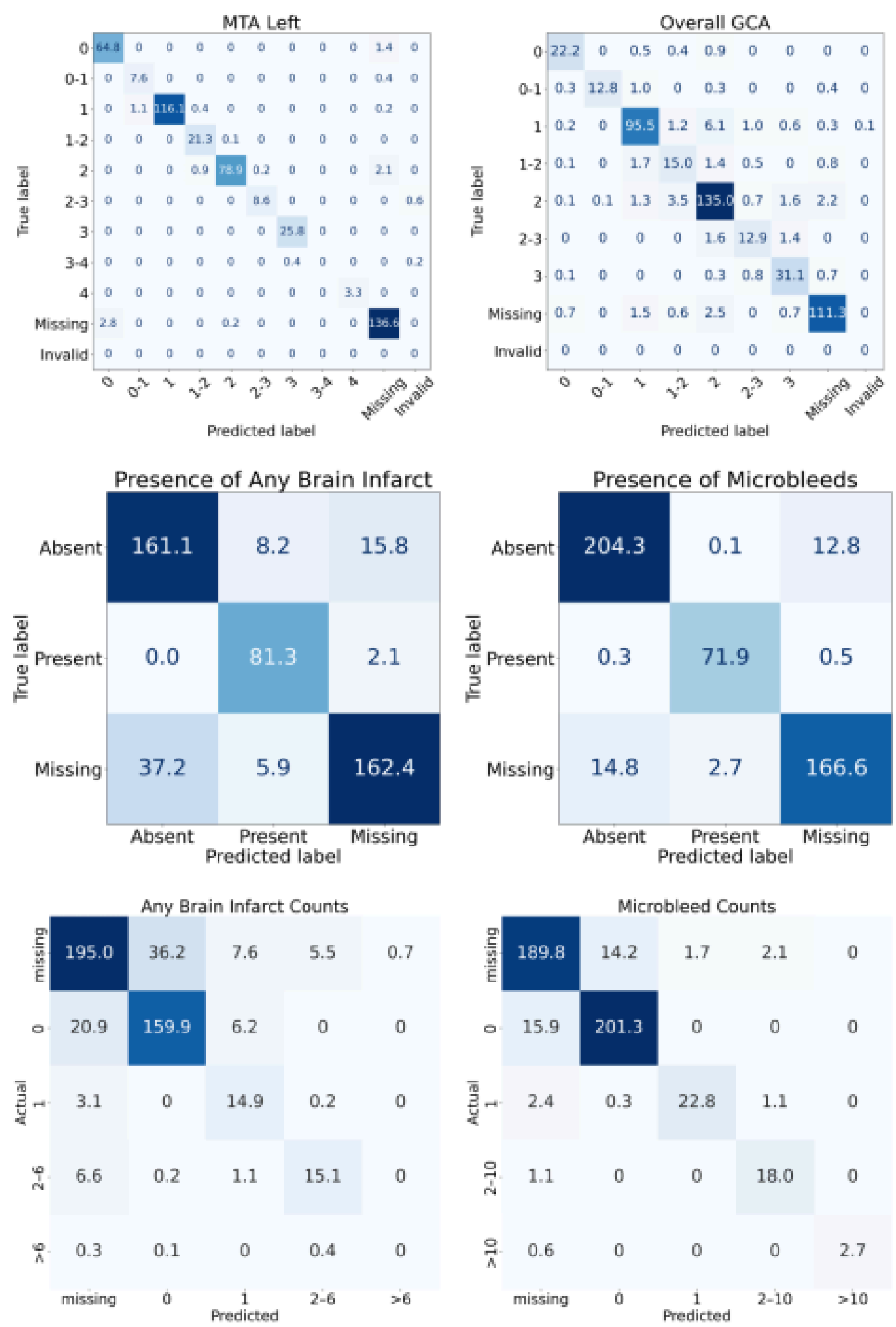

Figure 2: Confusion matrices normalized over the ten test sets of each random split obtained using few-shot prompting with structural similarity-based selection for the variables MTA Left, GCA Overall, Presence of Any Brain Infarcts, Presence of Microbleeds, Brain Infarct Counts and Microbleeds Counts.

## Discussion

This study evaluated the open-weight LLM LLaMA 3.1 for extracting structured data from Dutch radiology reports in a tertiary dementia clinic. We first assessed model performance compared to two human raters using 100 consensus-annotated reports. We then used the full dataset of 947 reports to examine the impact of input language and evaluate few-shot prompting with different example selection strategies as a resource-efficient alternative to fine-tuning.

LLaMA 3.1 demonstrated promising zero-shot capabilities in extracting 24 variables from Dutch brain MRI reports, performing overall comparably to human raters. We observed that the model shows excellent performance on simple variables (e.g. standardized visual scales). Performance was lower when extraction concerns specific locations in the brain (e.g. lobar GCAs) or specific lesion subtypes. Numerical variables pose additional difficulties, often demanding arithmetic operations (e.g., summing numbers of infarcts across different locations). Performance was also lower for variables that were less frequently reported by radiologists. Most variables showed moderate to near-perfect IRR, with only a few displaying poor agreement. Stratified analysis (Appendix E) found no link between agreement level and model performance, likely reflecting the sensitivity of IRR to class imbalance in rare findings such as superficial siderosis. However, lower agreement may introduce noise into the reference standard, so results for individual variables with weak agreement should be interpreted cautiously. Despite this, manual annotations remain the reference standard in radiological AI.

No significant performance gain was observed from translating the reports to English. Notably, the findings suggest that using the original Dutch reports yields more reliable results than using translations. A key challenge was handling local terminology. For example, the Dutch term "*splinterinfarcten*", referring to small, linear, streak-like cortical cerebellar infarcts, has no direct equivalent in the English literature and was therefore consistently mistranslated, leading to extraction errors. These infarcts are distinct from typical territorial cerebellar infarcts as their linear or radial configuration is consistent with the involvement of small penetrating vessels or distal emboli, aligning them more closely with small cortical infarcts and microinfarcts [34]. This morphological distinction motivated their separate analysis in our cohort. We addressed the mistranslation by including terminology-mapping in the translation prompt. This straightforward approach can be applied to other datasets but requires prior knowledge of local terms. Parameter-efficient fine-tuning methods could enable LLMs to learn local terminology and improve their domain knowledge directly from data, though current evidence suggests they underperform in limited-data settings [35]. We observed reduced performance in extracting free-text fields from translated reports. Upon examination, this reduction was primarily due to terminology discrepancies introduced during the translation of extracted content back into Dutch, which negatively impacted metric computations. Overall, the model showed a strong ability to interpret and process Dutch medical language, despite being primarily trained on English data.

We explored few-shot prompting as a resource and data-efficient alternative to fine-tuning, which is often impractical in clinical settings due to high computational and annotation costs. Our results show that few-shot prompting with curated example selection can significantly improve model performance. Variables that initially showed low performance in the zero-shot setting benefitted most. Data extraction required less than 30 seconds per report in our experimental setup using two NVIDIA H100 GPUs, corresponding to a throughput of approximately 120 reports per hour. While such hardware exceeds typical radiology workstation specifications, this configuration is feasible in centralized hospital IT environments, academic medical centers, and cloud-based infrastructures, where on-demand GPU instances are available at moderate hourly cost. The primary intended use case of large-scale registry generation and retrospective structuring of reports is inherently batch-oriented and does not require real-time processing; at the observed throughput, hospital-scale datasets can be processed in acceptable time frames. For prospective PACS/RIS integration, reports would be routed from the PACS/RIS to the extraction service, with structured outputs written back to the registry or electronic health record for radiologist review; automated structuring would occur in parallel with routine reporting and remain substantially faster than manual template completion. Several optimization strategies can further reduce hardware demands. Weight quantization (e.g., 8-bit or 4-bit) can lower the memory footprint of the 70B model to fit on a single GPU with modest accuracy degradation, and smaller model variants offer an alternative trade-off. LLaMA 3.1 8B can be deployed on a single consumer-grade GPU, however it achieved lower performance on most variables indicating that the accuracy gain from the 70B model is clinically meaningful (Appendix D). The 8B variant nonetheless remains a viable option for resource-constrained settings. Ongoing advances in model compression and inference hardware are expected to further narrow the gap toward seamless clinical deployment. Beyond the retrospective analysis reported in this work, real-world use would require a prospective validation on locally collected reports, periodic expert audits to detect drift, and confidence-based flagging of uncertain extractions for targeted review. Structured outputs should be treated as decision support rather than final records, with radiologist review incorporated for flagged cases and for variables with known weaker agreement. The traceability of each structured field back to specific text in the source report further supports efficient human verification. Any application of automated report structuring in clinical practice must also be considered in the context of regulatory and safety frameworks governing medical software. Tools that use patient data or influence clinical workflows may be subject to medical device regulations such as the European Union's Medical Device Regulation [36]. Using open-weight models that can be deployed on local institutional hardware avoids sending protected health information to external services and enables hospitals to maintain control over data governance and audit logs, potentially simplifying compliance with privacy and security standards. Nonetheless, rigorous validation, careful documentation of intended use, and adherence to relevant regulatory requirements will be necessary for any intended clinical deployment of LLM-based systems.

This study focused exclusively on extracting clinical variables from Dutch radiology reports in a tertiary memory clinic cohort using LLaMA 3.1. Although newer open-weight LLMs exist, LLaMA 3.1 remains competitive according to recent benchmarks[6]. Medical-specialized LLMs can offer advantages in more heterogeneous domains which require advanced clinical reasoning such as oncology, but evidence for their consistent benefit in structured information extraction is limited. For instance, Zhu et al. showed that an oncology-specific LLM outperformed LLaMA 3 in predicting cancer progression [37], yet Liu et al. [38] found that, without task-specific fine-tuning, general-purpose LLaMA 3.1 8B outperformed Llama-3-8B-UltraMedical on structuring radiology and pathology reports. In our cohort, reports contained structured elements such as visual rating scales, reducing the need for specialized medical knowledge and making a general-purpose model sufficient. A systematic comparison across newer general and medical-domain LLMs for this specific task remains an important direction for future work. The translation quality assessment was limited by its single-rater design and sample size (n=50), although the consistency of the observed differences and the downstream extraction results support the reported conclusions. Our study also lacked external validation, which limits its generalizability beyond the present cohort. Future research will extend to diverse and less standardized clinical domains to broaden the impact of these results. We employed few-shot prompting as a lightweight alternative to fine-tuning. More advanced approaches such as chain-of-thought prompting [14] and retrieval-augmented generation [39] may offer additional performance gains, though they were excluded from this study due to their computational and implementation demands. Direct comparison with proprietary models could further strengthen our conclusions; however, this was not feasible, as the dataset cannot be transferred to third-party servers. Moreover, the prompt used in this study was static throughout experiments. Future work may explore dynamic prompt generation techniques, such as automatic prompt optimization [40]. Finally, extending this approach to a multimodal setting by integrating both MRI scans and radiology reports as input to a vision-language model [41] could enable more comprehensive and accurate clinical information extraction.

[6] www.vellum.ai/open-llm-leaderboard (Update of 19 November 2025)

## Conclusion

This study demonstrates the potential of open-weight LLMs for extracting structured clinical information from radiology reports. Few-shot prompting with informed example selection significantly improves performance, approaching human-level accuracy for most variables. Although challenges persist for complex or ambiguous cases, the model's effectiveness in processing Dutch medical text underscores the promise of LLMs as scalable tools for post-hoc structuring of free-text reports. Future work should validate this approach across models, extend it to other clinical domains, and explore multimodal integration to enhance its applicability across different research settings.

# Appendix

## A. Prompt

The prompt used in this work along with the JSON template are presented below

***System part:***

You are a system designed to structure free-text radiology reports. Your task is to extract relevant information and organize it according to a specific JSON template. Follow these guidelines:

1. Follow the instructions given in the JSON template below for each variable.
2. Respond only with the structured report in JSON format. Do not include any explanations, commentary, or reasoning.
3. If the input text is not a valid radiology report, return an empty JSON object ('{}').

**JSON Template:**

{"Fazekas": "Return the Fazekas score. If Fazekas scores are given for both left and right, return the highest. If given as a range (e.g., 1-2), return the average (e.g., 1.5). If not mentioned, return 'missing'.",
"WMH_absent": "Return 1 if the report explicitly states absence of white matter hyperintensities (WMH), *without* mentioning a Fazekas score. Otherwise, return 0.",
"MTA_L": "Return the medial temporal atrophy (MTA) score for the left side. If a range is provided, return the average. If not mentioned, return 'missing'.",
"MTA_R": "Return the medial temporal atrophy (MTA) score for the right side. If a range is provided, return the average. If not mentioned, return 'missing'.",
"hippocampus_normal": "Return 1 if the hippocampus is explicitly described as normal (e.g., 'normal hippocampal volume'), *without* mentioning an MTA score. Otherwise, return 0.",
"GCA_front": "Return the global cortical atrophy (GCA) score of the frontal lobe. If a range is given, return the average. If only a generalized GCA score is reported, use that. If not mentioned, return 'missing'.",
"GCA_temp": "Return the GCA score of the temporal lobe. If a range is given, return the average. If only a generalized GCA score is reported, use that. If not mentioned, return 'missing'.",
"GCA_occip": "Return the GCA score of the occipital lobe. If a range is given, return the average. If only a generalized GCA score is reported, use that. If not mentioned, return 'missing'.",
"GCA_pari": "Return the GCA score of the parietal lobe. If a range is given, return the average. If only a generalized GCA score is reported, use that. If not mentioned, return 'missing'.",
"GCA_overall": "Return the overall/generalized GCA score. If only lobar/partial GCA scores are provided, return the highest partial GCA score. If a range is given, return the average. If not mentioned, return 'missing'.",
"atrophy_absent": "Return 1 if the report explicitly states absence of atrophy *without* providing any GCA score. Otherwise, return 0.",
"signal_intensity": "Return 1 if the report mentions normal signal intensity of the grey and white matter (or equivalent). Otherwise, return 0.",
"microbleeds": "Return 1 if microbleeds are present, 0 if explicitly absent (no microbleeds). If not mentioned, return 'missing'.",
"microbleeds_n": "Return the total number of microbleeds. If not mentioned, return 'missing'.",
"microbleeds_location": "Return the reported location(s) of microbleeds. If not mentioned, return 'missing'.",
"infarct_cortical": "Return 1 if cortical infarcts are present, 0 if explicitly absent. If not mentioned, return 'missing'.",
"infarct_cortical_n": "Return the total number of cortical infarcts. If not mentioned, return 'missing'.",

"infarct_lacunar": "Return 1 if lacunar infarcts (lacunes) are present, 0 if explicitly absent. If not mentioned, return 'missing'.",
"infarct_lacunar_n": "Return the total number of lacunar infarcts. If not mentioned, return 'missing'.",
"infarct_cerebellar": "Return 1 if cerebellar infarcts (of any type) are present, 0 if explicitly absent. If not mentioned, return 'missing'.",
"infarct_cerebellar_n": "Return the total number of cerebellar infarcts. If not mentioned, return 'missing'.",
"infarct_splinter": "Return 1 if splinter infarcts are present, 0 if explicitly absent. If not mentioned, return 'missing'.",
"infarct_any": "Return 1 if any type of infarct is present, 0 if explicitly absent (no infarcts). If not mentioned, return 'missing'.",
"infarct_any_n": "Return the total number of infarcts (of any type) if specified. If not mentioned, return 'missing'.",
"infarct_location": "Return the reported location(s) of infarcts. If not mentioned, return 'missing'.",
"DWI": "Return 1 if diffusion-weighted imagine (DWI) shows diffusion restriction. Return 0 if DWI is explicitly normal (no diffusion restriction). If not mentioned, return 'missing'.",
"SWI": "Return 1 if susceptibility-weighted imaging (SWI) shows abnormalities. Return 0 if SWI explicitly stated as normal. If not mentioned, return 'missing'.",
"siderosis": "Return 1 if superficial siderosis is mentioned. Otherwise, return 0.",
"artifact": "Return 1 if imaging artifacts (e.g., motion) affecting interpretation are mentioned. Otherwise, return 0.",
"quantib": "Indicate 1 if Quantib segmentation (or quantitative evaluation) is mentioned. otherwise return 0."

***User part:***

**Radiology Report:**

{the report to be structured}

## B. Variable definitions and distributions

Table S1: Variable definitions and distributions for the total dataset and the 100 double-annotated reports.

| Variable | Definition | Distribution | | |
|---|---|---|---|---|
| Categorical variables | | Value | N: total dataset (n=947) | N: double-annotated subset (n=100) |
| MTA Left | Medial temporal atrophy score of the left side | missing | 285 | 33 |
| | | 0 | 135 | 10 |
| | | 0-1 | 16 | 2 |
| | | 1 | 232 | 22 |
| | | 1-2 | 43 | 2 |
| | | 2 | 160 | 18 |
| | | 2-3 | 16 | 1 |
| | | 3 | 52 | 12 |
| | | 3-4 | 1 | 0 |
| | | 4 | 7 | 0 |
| MTA Right | Medial temporal atrophy score of the right side | missing | 284 | 33 |
| | | 0 | 132 | 10 |
| | | 0-1 | 19 | 2 |
| | | 1 | 228 | 25 |
| | | 1-2 | 43 | 1 |
| | | 2 | 166 | 20 |
| | | 2-3 | 21 | 2 |
| | | 3 | 47 | 7 |
| | | 3-4 | 3 | 0 |
| | | 4 | 4 | 0 |
| Normal hippocampus | Binary indicator set to 1 if the report only mentions normal hippocampal volume without providing a specific MTA score. | 0 | 856 | 89 |
| | | 1 | 91 | 11 |
| GCA frontal | Frontal lobe global cortical atrophy (GCA) score. If a frontal lobe-specific score is unavailable, the whole-brain GCA score may be used instead. | missing | 282 | 35 |
| | | 0 | 81 | 8 |
| | | 0-1 | 29 | 3 |
| | | 1 | 258 | 20 |
| | | 1-2 | 52 | 4 |
| | | 2 | 197 | 24 |
| | | 2-3 | 22 | 1 |
| | | 3 | 26 | 5 |
| GCA temporal | Global cortical atrophy score of the temporal lobe. If a temporal lobe-specific score is unavailable, the whole-brain GCA score may be used instead. | missing | 305 | 36 |
| | | 0 | 98 | 7 |
| | | 0-1 | 31 | 3 |
| | | 1 | 269 | 23 |
| | | 1-2 | 46 | 4 |
| | | 2 | 145 | 21 |
| | | 2-3 | 19 | 2 |
| | | 3 | 34 | 4 |
| GCA occipital | Global cortical atrophy score of the occipital lobe. If an occipital lobe-specific score is unavailable, the whole-brain GCA score may be used instead. | missing | 361 | 41 |
| | | 0 | 134 | 8 |
| | | 0-1 | 34 | 4 |
| | | 1 | 275 | 23 |
| | | 1-2 | 35 | 3 |
| | | 2 | 89 | 17 |
| | | 2-3 | 7 | 1 |
| | | 3 | 12 | 3 |
| GCA parietal | Global cortical atrophy score of the parietal lobe. If a parietal lobe-specific score is unavailable, the whole-brain GCA score may be used instead. | missing | 256 | 34 |
| | | 0 | 58 | 7 |
| | | 0-1 | 28 | 3 |
| | | 1 | 208 | 15 |
| | | 1-2 | 47 | 6 |
| | | 2 | 284 | 29 |
| | | 2-3 | 30 | 1 |
| | | 3 | 36 | 5 |
| GCA overall | Global cortical atrophy score of the whole brain. If the score is given separately for different regions of the brain without specifying a global score, the largest of the partial values is reported. | missing | 239 | 31 |
| | | 0 | 48 | 7 |
| | | 0-1 | 29 | 3 |
| | | 1 | 208 | 22 |
| | | 1-2 | 39 | 2 |
| | | 2 | 285 | 26 |
| | | 2-3 | 33 | 3 |
| | | 3 | 66 | 6 |
| Absence of atrophy | Binary indicator set to 1 if the report only mentions no atrophy (or similar) without providing a specific GCA score | 0 | 901 | 93 |
| | | 1 | 46 | 7 |
| Fazekas | Fazekas score as mentioned in the report. If the score is given separately for the left and right side, the largest of the two values is reported. | missing | 270 | 31 |
| | | 0 | 163 | 9 |
| | | 0-1 | 12 | 2 |
| | | 1 | 274 | 30 |
| | | 1-2 | 24 | 2 |
| | | 2 | 120 | 12 |
| | | 2-3 | 14 | 4 |
| | | 3 | 70 | 10 |
| Absence of WMH | Binary indicator set to 1 if the report mentions the absence of white matter hyperintensity without providing a specific Fazekas score. | 0 | 913 | 95 |
| | | 1 | 34 | 5 |
| Normal signal intensity | Binary indicator set to 1 if the report mentions a normal signal intensity of the white and gray matter (or similar). | 0 | 876 | 92 |
| | | 1 | 71 | 8 |
| Presence of cortical infarcts | Whether the report mentions the presence or absence of cortical infarcts in the brain. If cortical infarcts are not mentioned the variable is considered missing. | missing | 447 | 64 |
| | | absent | 465 | 32 |
| | | present | 35 | 4 |
| Presence of lacunes | Whether the report mentions the presence or absence of lacunes in the brain. If cortical infarcts are not mentioned the variable is considered missing. | missing | 389 | 51 |
| | | absent | 423 | 32 |
| | | present | 135 | 17 |
| Presence of cerebellar infarcts | Whether the report mentions the presence or absence of cerebellar infarcts in the brain. If cerebellar infarcts are not mentioned the variable is considered missing. | missing | 504 | 62 |
| | | absent | 396 | 28 |
| | | present | 47 | 10 |
| Presence of splinter infarcts | Whether the report mentions the presence or absence of splinter infarcts (small oblique cerebellar infarcts) in the brain. If splinter infarcts are not mentioned the variable is considered missing. | missing | 529 | 69 |
| | | absent | 402 | 28 |
| | | present | 16 | 3 |
| Presence of any infarcts | Whether the report mentions the presence or absence of any type of infarcts in the brain. If infarcts are not mentioned the variable is considered missing. | missing | 412 | 48 |
| | | absent | 364 | 29 |
| | | present | 171 | 23 |
| Presence of microbleeds | Whether the report mentions the presence or absence of microbleeds in the brain. If microbleeds are not mentioned the variable is considered missing. | missing | 369 | 50 |
| | | absent | 430 | 31 |
| | | present | 148 | 19 |
| SWI abnormalities | Whether the report mentions the presence or absence of susceptibility imaging abnormalities. If the SWI is not mentioned the variable is considered missing. | missing | 848 | 80 |
| | | absent | 68 | 11 |
| | | present | 31 | 9 |
| DWI abnormalities | Whether the report mentions the presence or absence of diffusion restriction on the DWI sequence. If the DWI is not mentioned the variable is considered missing. | missing | 387 | 52 |
| | | absent | 527 | 44 |
| | | present | 33 | 4 |
| Siderosis | Whether the report mentions the presence or absence of superficial siderosis. If superficial siderosis is not mentioned it is considered absent. | absent | 930 | 99 |
| | | present | 17 | 1 |
| Quantitative evaluation | Indicator set to 1 if the report mentions the use of quantitative evaluation using the Quantib software. | 0 | 686 | 98 |
| | | 1 | 261 | 2 |
| Artifacts | Indicator set to 1 if the report mentions the presence of imaging artifacts (e.g. motion) hindering interpretation. | 0 | 911 | 91 |
| | | 1 | 36 | 9 |
| Numerical Variables | | Statistic | Total dataset | Double-annotated subset |
| Cortical infarcts | Number of cortical infarcts. If not mentioned the value is set to -1. | Number missing | 459 | 64 |
| | | Number = 0 | 465 | 32 |
| | | Mean of non-missing | 0.06 | 0.14 |
| | | STD of non-missing | 0.32 | 0.42 |
| Lacunes | Number of lacunes. If not mentioned the value is set to -1. | Number missing | 429 | 57 |
| | | Number = 0 | 422 | 32 |
| | | Mean of non-missing | 0.47 | 0.60 |
| | | STD of non-missing | 1.24 | 1.37 |

| Cerebellar infarcts | Number of cerebellar infarcts. If not mentioned the value is set to -1. | Number missing | 520 | 65 |
|---|---|---|---|---|
| | | Number = 0 | 398 | 28 |
| | | Mean of non-missing | 0.11 | 0.31 |
| | | STD of non-missing | 0.49 | 0.90 |
| All infarcts | Number of all infarcts. If not mentioned the value is set to -1. | Number missing | 495 | 55 |
| | | Number = 0 | 368 | 29 |
| | | Mean of non-missing | 0.43 | 0.78 |
| | | STD of non-missing | 1.23 | 1.41 |
| Microbleeds | Number of microbleeds. If not mentioned the value is set to -1. | Number missing | 417 | 59 |
| | | Number = 0 | 430 | 31 |
| | | Mean of non-missing | 0.68 | 1.85 |
| | | STD of non-missing | 3.61 | 7.26 |
| Location variables (text fields) | | Statistic | Total dataset | Double-annotated subset |
| Infarcts | Text description of the locations of all infarcts. | N non-missing | 165 | 23 |
| Microbleeds | Text description of the locations of all microbleeds. | N non-missing | 146 | 19 |

N: number of samples

## C. Results for quality control variables

Table S2: Quantitative performance of the LLaMA 3.1 against two human raters and the inter-rater reliability of the two raters on the QC variables. A consensus rating was used as a reference standard.

| | **LLaMA 3.1** | **Annotator 1** | **Annotator 2** | **Inter-rater reliability** |
|---|---|---|---|---|
| **Categorical Variables** | | **Balanced Accuracy (%)** | | **Cohen's κ** |
| **QC Variables** | | | | |
| Normal hippocampus | 86 | 99* | 91* | 0.84 |
| Absence of atrophy | 68 | 92 | 100 | 0.78 |
| Absence of WMH | 77 | 99 | 70 | 0.43 |
| Normal signal intensity | 82 | 81 ⁑ | 92* | 0.43 |
| Quantitative evaluation | 75 | 75 | 100 | 0.66 |
| Presence of artifacts | 97 | 89 | 78 | 0.47 |

ICC: Intra-class correlation; (*) The rater significantly outperforms the model, and (⁑) The model significantly outperforms the rater, both based on the McNemar test ($p < 0.05$).

Table S3: Zero-shot information extraction performance on original Dutch reports vs. English translations by LLaMA 3.1 and OPUS-MT on the QC variables.

| | **Dutch report** | **LLaMA 3.1 Translation** | **OPUS-MT Translation** |
|---|---|---|---|
| **Categorical Variables (%)** | | **Balanced Accuracy (%)** | |
| **QC Variables** | | | |
| Normal hippocampus | 84 (81, 87) | 82 (80, 85) | 78 (74, 83) |
| Absence of atrophy | 74 (65, 82) | 76 (68, 84) | 72 (65, 79) |
| Absence of WMH | 85 (77, 94) | 82 (74, 91) | 77 (69, 84) |
| Normal signal intensity | 77 (75, 79) | 78 (75, 80) | 74 (69, 79) |
| Quantitative evaluation | 88 (86, 91) | **96 (95, 97)** | 70 (65, 75) |
| Presence of artifacts | 97 (96, 98) | 97 (96, 97) | 91 (84, 99) |

Scores are reported as: value (95%-CI). **Bold** means translation provides significantly better performance. Underlined means translation provides significantly worse performance.

Table S4: Few-shot information extraction performance using fixed examples, randomly selected examples, structural (Jaccard) similarity-based examples and semantic similarity-based examples.

| **Selection method** | **Zero-shot** | **Random** | **Fixed** | **Structural Similarity** | **Semantic Similarity** |
|---|---|---|---|---|---|
| **Categorical Variables** | **Balanced Accuracy (%)** | | | | |
| **QC Variables** | | | | | |
| Normal hippocampus | 84 (81, 87) | 82 (79, 86) | 84 (74, 94) | 84 (79, 89) | 85 (82, 88) |
| Absence of atrophy | 74 (65, 82) | 77 (69, 86) | 78 (69, 86) | 79 (69, 90) | 79 (71, 86) |
| Absence of WMH | 85 (77, 94) | 79 (71, 86) | 80 (63, 98) | 84 (74, 95) | 82 (68, 96) |
| Normal signal intensity | 77 (75, 79) | 77 (69, 86) | 81 (77, 84) | 77 (70, 85) | 79 (72, 87) |
| Quantitative evaluation | 88 (86, 91) | 92 (89, 94) | 86 (70, 100) | 95 (90, 100) | 94 (90, 97) |
| Presence of artifacts | 97 (96, 98) | 97 (94, 100) | 96 (92, 100) | 97 (97, 98) | 96 (92, 99) |

## D. Comparison between LLaMA 3.1 70B and LLaMA 3.1 8B

Table S5: Zero-shot performance comparison between the 70B and 8B parameters versions of LLaMA 3.1

| | **LLaMA 3.1 70B** | **LLaMA 3.1 8B** |
|---|---|---|
| **Categorical Variables** | **Balanced Accuracy** | |
| **Atrophy Measures** | | |
| MTA left | 90 (77, 100) | 81 (70, 93) |
| MTA right | **96 (94, 99)** | 86 (80, 93) |
| GCA overall | **87 (83, 91)** | 71 (65, 76) |
| GCA frontal | **80 (73, 87)** | 65 (59, 71) |
| GCA temporal | **73 (69, 78)** | 65 (62, 68) |
| GCA occipital | 58 (47, 69) | 41 (34, 48) |
| GCA parietal | **77 (70, 84)** | 59 (53, 64) |
| **Vascular Markers** | | |
| Fazekas score | **94 (93, 96)** | 82 (75, 89) |
| Any brain infarcts | 82 (80, 84) | 85 (82, 87) |
| Cortical infarcts | 75 (69, 81) | 78 (74, 83) |
| Lacunes | 83 (80, 86) | 89 (86, 91) |
| Cerebellar infarcts | 69 (66, 71) | **80 (74, 86)** |
| Splinter infarcts | 66 (64, 67) | 77 (69, 85) |
| Microbleeds | 93 (92, 95) | 91 (90, 92) |
| SWI abnormalities | 70 (61, 79) | 59 (48, 70) |
| Superficial siderosis | 97 (85, 100) | 97 (85, 100) |
| DWI abnormalities | 89 (81, 96) | 86 (79, 93) |
| **Variables for Quality Control S** | | |
| Normal hippocampus | **84 (81, 87)** | 77 (76, 79) |
| Absence of atrophy | 74 (65, 82) | 78 (69, 86) |
| Absence of WMH | **85 (77, 94)** | 70 (65, 75) |
| Normal signal intensity | **77 (75, 79)** | 61 (60, 62) |
| Quantitative evaluation | 88 (86, 91) | 93 (90, 97) |
| Presence of artifacts | 97 (96, 98) | 94 (88, 100) |
| **Numerical Variables** | **Accuracy [Mean Absolute Error]** | |
| Any brain infract | **66 (63, 68)** [0.04] | 54 (50, 59) [0.48] |
| Cortical infarcts | **61 (58, 64)** [0.01] | 51 (47, 56) [0.11] |
| Lacunes | **69 (66, 72)** [0.10] | 52 (47, 57) [0.32] |
| Cerebellar infarcts | 61 (58, 64) [0.03] | 55 (52, 59) [0.41] |
| Microbleeds | **80 (78, 82)** [0.00] | 54 (51, 57) [0.03] |
| **Text Variables** | **Text Similarity** | |
| Location of infarcts | **0.95 (0.95, 0.95)** | 0.94 (0.93, 0.94) |
| Location of microbleeds | 0.95 (0.95, 0.96) | 0.94 (0.94, 0.95) |

## E. Impact of inter-rater disagreement on the reliability of single-rater annotations

### E.1 Bootstrapping Analysis

We assessed the reliability of using individual rater annotations versus consensus annotations as a reference for evaluating model performance.  To estimate variability and generate confidence intervals, we performed a bootstrap experiment with 500 iterations. In each iteration, reports were sampled with replacement to create a bootstrap sample of the same size as the original dataset. Model performance was computed in two ways for each sample:

5. Consensus reference: the model predictions were compared to the consensus annotations of the bootstrap-sampled reports.

6. Single-rater reference: for each bootstrap-sampled report, one of the two raters (R1 or R2) was selected randomly and independently, and the model prediction was compared to that rater's annotation.

This process was repeated across all iterations, producing distributions of performance metrics for each variable. The mean performance and 95% confidence intervals were then calculated across bootstrap iterations. The results are summarized in Fig. S1.

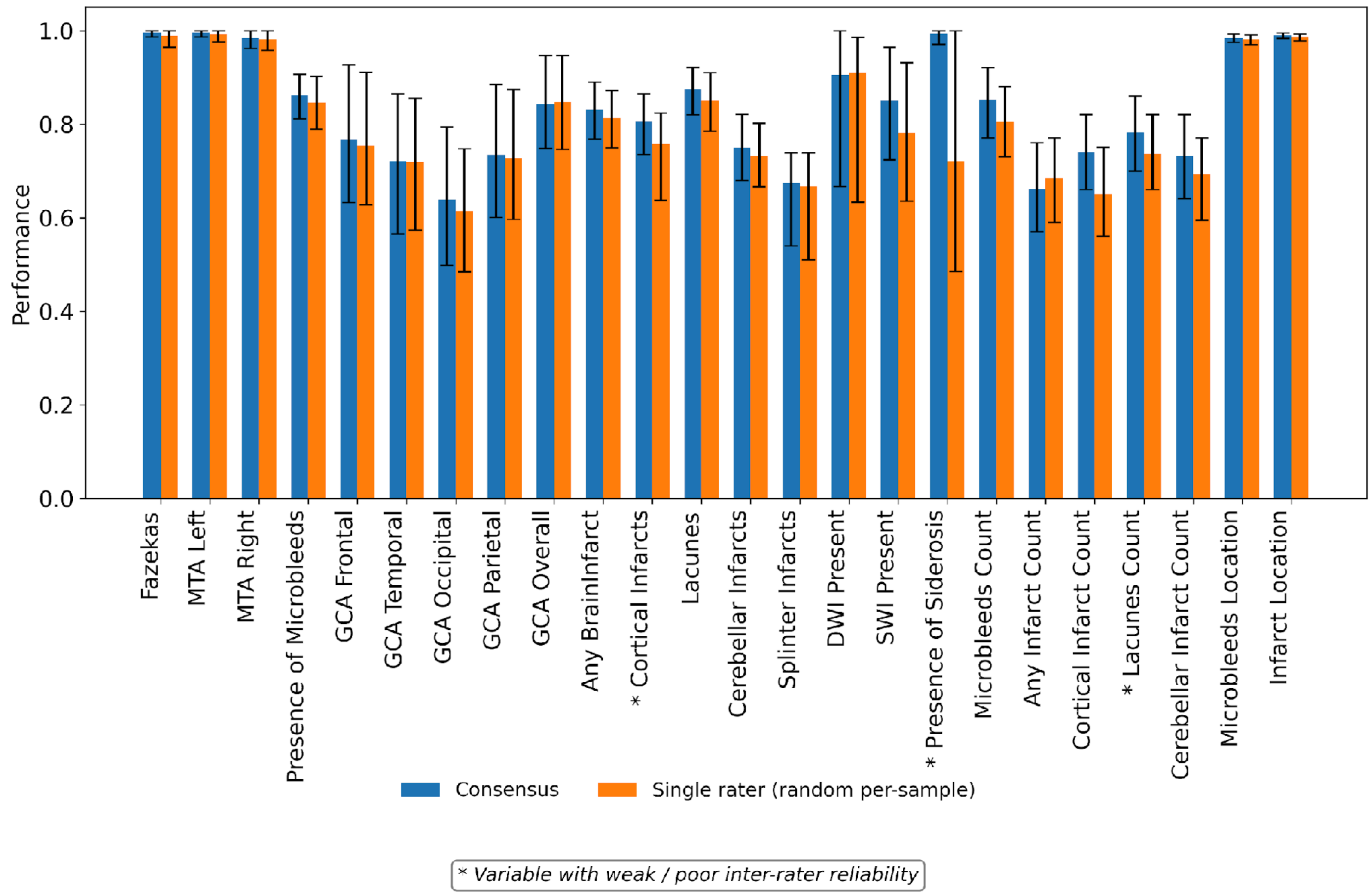

Figure S1: Comparison of performance measures obtained using the consensus annotations and single-rater annotations on the 100 double-annotated reports. Error bars show the 95% confidence intervals computed over 500 bootstrapped iterations.

Model performance was generally higher when evaluated against the consensus annotations than when evaluated against a single-rater reference chosen randomly per sample. For most variables, the mean performance differences were modest, and the 95% confidence intervals largely overlapped, indicating that the two evaluation approaches are broadly comparable.

For variables with weaker inter-rater reliability, the mean difference in performance between the consensus rating and the single-rater reference was similar to or larger than for other variables, reflecting the influence of rater variability. Categorical variables such as Fazekas scores and MTA ratings exhibited near-perfect performance in both conditions. Variables with lower prevalence or more subjective assessment (e.g., cortical infarcts, splinter infarcts, presence of superficial siderosis) showed wider confidence intervals with single-rater evaluation. Overall, these results indicate that while using consensus annotations provides a slightly more stable reference, evaluation with randomly chosen single-rater annotations yields comparable performance estimates across most variables.

## E.2 Sensitivity Analysis by Agreement Level Stratification

We assessed how inter-rater agreement relates to model performance across the full evaluation set (n = 947). Variables were stratified into three agreement bins. Within each bin, we report the number of variables, mean IRR, and the mean and standard deviation of model performance (Table S6). To quantify the relationship between IRR and model performance, we computed Spearman's rank correlation between the IRR value and the performance metric across variables. Because variable types use different performance and IRR metrics, the primary correlation was restricted to categorical variables (n = 17), which share a single IRR metric (Cohen's κ) and a single performance metric (balanced accuracy). A 95% confidence interval for Spearman's ρ was obtained via percentile bootstrap with 1000 resamples. Numerical variables (n = 5) are reported descriptively given the small sample size, and text variables (n = 2) are reported for completeness. The relationship across all 24 variables is visualized in Fig. S2.

Table S6: Model performance stratified by inter-rater agreement level. Bins are defined by field-standard thresholds for each IRR metric (Cohen's κ, ICC, or text similarity). Performance refers to balanced accuracy for categorical variables, accuracy for numerical variables, and semantic similarity for text variables.

| Variable type | Bin | N variables | Mean IRR | Performance (mean ± std) |
|---|---|---|---|---|
| **Categorical** | High (κ ≥ 0.80) | 10 | 0.93 | 82.0 ± 11.1 |
| | Moderate (0.60 ≤ κ < 0.80) | 5 | 0.72 | 77.4 ± 12.6 |
| | Low (κ < 0.60) | 2 | 0.49 | 86.0 ± 15.6 |
| **Numerical** | High (ICC ≥ 0.75) | 2 | 0.90 | 70.5 ± 13.4 |
| | Moderate (0.50 ≤ ICC < 0.75) | 2 | 0.68 | 63.5 ± 3.5 |
| | Low (ICC < 0.50) | 1 | 0.45 | 69.0 |
| **Text field** | High (similarity ≥ 0.90) | 2 | 0.94 | 95.0 ± 0.0 |

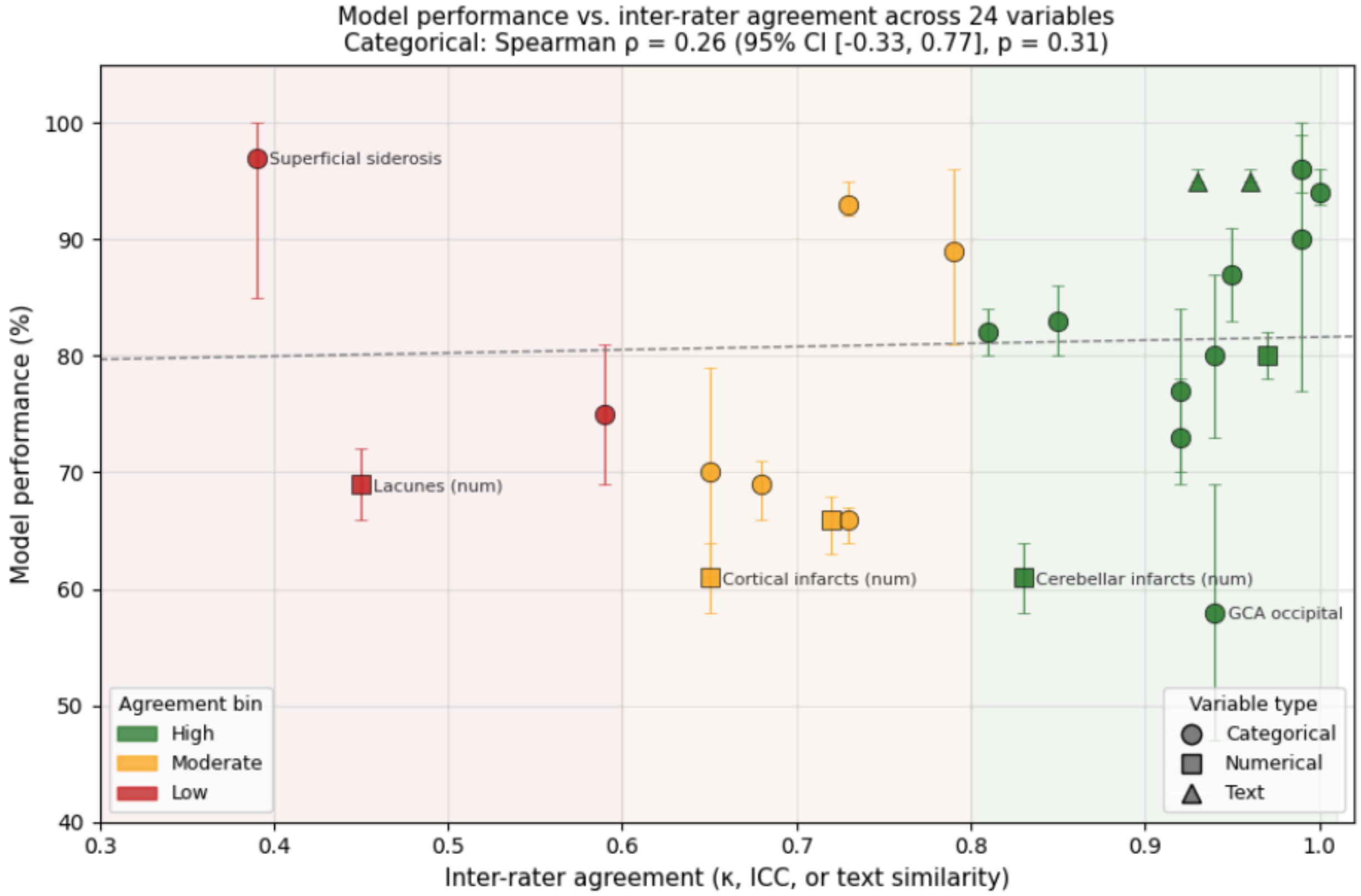

Fig S2: Model performance versus inter-rater agreement across all 24 variables. Marker shape indicates variable type (categorical, numerical, text), and color indicates agreement bin (High, Moderate, Low) based on the categorical variables. Error bars show 95% confidence intervals on performance. The dashed line is a linear fit to the categorical variables, shown for visual reference.

Across categorical variables, we found no statistically significant correlation between Cohen's κ and balanced accuracy (Spearman ρ = 0.26, 95% CI [−0.34, 0.77], p = 0.31). Mean balanced accuracy was comparable across the three agreement bins, and performance in the Low-agreement bin was not lower than in the High-agreement bin. This pattern reflects the sensitivity of κ to label skew: for rare findings such as superficial siderosis (only one of the 100 double-annotated reports labeled positive), κ is negatively affected by class imbalance despite high raw agreement, and the model achieves high balanced accuracy. For numerical variables, the Spearman correlation between ICC and accuracy was similarly non-significant (ρ = 0.21, p = 0.74, n = 5), although this estimate should be interpreted as indicative given the small number of variables. Both text variables fell in the High-agreement bin and achieved high similarity scores.
Overall, these results indicate that model performance is not systematically degraded on variables with weaker inter-rater agreement. Together with the bootstrap analysis above, this suggests that annotation uncertainty, while a legitimate concern for individual low-agreement variables, does not broadly compromise the reported performance estimates.

## F. Results for spliter infarcts-specific translation

Table S7: Extraction performance obtained using the general LLaMA 3.1 translation and the LLaMA 3.1 translation with specific instructions for splinter infarcts.

| | General translation | Splinter infarcts-specific translation |
|---|---|---|
| **Categorical Variables** | **Balanced Accuracy** | |
| **Atrophy Measures** | | |
| MTA left | 90 (77, 100) | 91 (79, 100) |
| MTA right | 96 (94, 99) | 95 (88, 100) |
| GCA overall | 88 (84, 91) | 88 (84, 91) |
| GCA frontal | 81 (75, 87) | 81 (75, 87) |
| GCA temporal | 78 (73, 82) | 76 (71, 80) |
| GCA occipital | 61 (51, 72) | 60 (51, 69) |
| GCA parietal | 78 (73, 83) | 79 (74, 83) |
| **Vascular Markers** | | |
| Fazekas score | 94 (93, 96) | 94 (92, 96) |
| Any brain infarcts | 81 (79, 84) | 82 (80, 85) |
| Cortical infarcts | 74 (67, 81) | 75 (69, 81) |
| Lacunes | 85 (82, 88) | 84 (79, 88) |
| Cerebellar infarcts | 68 (65, 71) | 69 (65, 73) |
| *Splinter infarcts* | *37 (31, 44)* | *65 (59, 70)* |
| Microbleeds | 94 (92, 95) | 94 (93, 96) |
| SWI abnormalities | 75 (68, 82) | 73 (64, 82) |
| Presence of siderosis | 94 (80, 100) | 94 (80, 100) |
| DWI abnormalities | 85 (77, 94) | 85 (76, 94) |
| **Numerical Variables** | **Accuracy [Mean Absolute Error]** | |
| Cortical infarcts | 62 (59, 66) [0.01] | 62 (59, 65) [0.05] |
| Lacunes | 75 (71, 78) [0.08] | 70 (67, 74) [0.07] |
| Cerebellar infarcts | 62 (58, 67) [0.03] | 61 (56, 65) [0.05] |
| All infarcts | 67 (63, 70) [0.04] | 65 (62, 69) [0.07] |
| Microbleeds | 81 (78, 84) [0.00] | 77 (75, 80) [0.00] |
| **Text Variables** | **Text Similarity** | |
| Location of infarcts | 0.93 (0.92, 0.93) | 0.93 (0.92, 0.93) |
| Location of microbleeds | 0.91 (0.90, 0.91) | 0.91 (0.90, 0.92) |

## G. Evaluation using additional metrics

Table S8: Performance of the LLaMA 3.1 and two human raters against the consensus annotations measured using F1-Score for categorical and Spearman's correlation for numerical variables

| | LLaMA 3.1 vs Consensus | Rater 1 vs Consensus | Rater 2 vs Consensus |
|---|---|---|---|
| **Categorical Variables** | **F1-Score** | | |
| **Atrophy Measures** | | | |
| MTA left | 1.0 | 0.99 | 1.0 |
| MTA right | 0.98 | 0.99 | 1.0 |
| GCA overall | 0.76 | 0.94 | 1.0 |
| GCA frontal | 0.76 | 0.96 | 0.92 |
| GCA temporal | 0.79 | 0.96 | 0.94 |
| GCA occipital | 0.67 | 0.96 | 0.90 |
| GCA parietal | 0.74 | 0.97 | 0.92 |
| **Vascular Markers** | | | |
| Fazekas score | 0.97 | 0.99 | 1.0 |
| Any brain infarct | 0.80 | 0.92 | 0.91 |
| Cortical infarcts | 0.71 | 0.94 | 0.81 |
| Lacunes | 0.84 | 0.98 | 0.94 |
| Cerebellar infarcts | 0.74 | 0.88 | 0.91 |
| Splinter infarcts | 0.68 | 0.88 | 0.94 |
| Microbleeds | 0.82 | 0.90 | 0.95 |
| SWI abnormalities | 0.87 | 0.92 | 0.82 |
| Superficial siderosis | 0.83 | 0.69 | 1.0 |
| DWI abnormalities | 0.94 | 0.78 | 0.95 |
| **Numerical Variables** | **Spearman's Correlation [number of extracted entries]** | | |
| Any brain infarct | 1.00 [13] | 0.94 [34] | 0.93 [42] |
| Cortical infarcts | NA [11] | 0.82 [35] | 1.00 [35] |
| Lacunes | 0.93 [25] | 1.00 [41] | 1.00 [42] |
| Cerebellar infarcts | 1.00 [11] | 0.89 [32] | 1.00 [34] |
| Microbleeds | 1.00 [27] | 0.95 [37] | 1.00 [39] |

Table S9: Zero-shot information extraction performance on original Dutch reports vs. English translations by LLaMA 3.1 and OPUS-MT measured using F1-score for categorical and Spearman's correlation for numerical variables.

| | **Dutch report** | **LLaMA 3.1 Translation** | **OPUS-MT Translation** |
|---|---|---|---|
| **Categorical Variables (%)** | **F1-Score** | | |
| **Atrophy Measures** | | | |
| MTA left | 0.83 (0.66, 1.0) | 0.81 (0.70, 0.93) | 0.73 (0.56, 0.90) |
| MTA right | 0.91 (0.80, 1.0) | 0.88 (0.86, 0.90) | 0.79 (0.67, 0.91) |
| GCA overall | 0.79 (0.76, 0.82) | 0.79 (0.77, 0.81) | 0.73 (0.71, 0.75) |
| GCA frontal | 0.75 (0.64, 0.86) | 0.77 (0.66, 0.87) | 0.69 (0.57, 0.80) |
| GCA temporal | 0.71 (0.67, 0.75) | 0.76 (0.65, 0.88) | 0.64 (0.61, 0.68) |
| GCA occipital | 0.58 (0.47, 0.68) | 0.67 (0.55, 0.79) | 0.53 (0.46, 0.61) |
| GCA parietal | 0.72 (0.67, 0.77) | 0.73 (0.69, 0.77) | 0.65 (0.55, 0.76) |
| **Vascular Markers** | | | |
| Fazekas score | 0.90 (0.86, 0.93) | 0.91 (0.87, 0.95) | 0.77 (0.63, 0.90) |
| Any brain infarct | 0.80 (0.78, 0.82) | 0.80 (0.78, 0.82) | 0.67 (0.65, 0.70) |
| Cortical infarcts | 0.71 (0.65, 0.78) | 0.70 (0.65, 0.75) | 0.60 (0.55, 0.65) |
| Lacunes | 0.83 (0.80, 0.86) | 0.84 (0.81, 0.88) | 0.66 (0.61, 0.70) |
| Cerebellar infarcts | 0.65 (0.61, 0.69) | 0.64 (0.60, 0.67) | 0.59 (0.54, 0.64) |
| Splinter infarcts | 0.61 (0.56, 0.65) | 0.36 (0.26, 0.47) | 0.53 (0.43, 0.63) |
| Microbleeds | 0.93 (0.91, 0.95) | 0.94 (0.92, 0.96) | 0.65 (0.61, 0.69) |
| SWI abnormalities | 0.73 (0.64, 0.81) | 0.78 (0.72, 0.84) | 0.61 (0.52, 0.70) |
| Presence of siderosis | 0.95 (0.87, 1.0) | 0.97 (0.89, 1.0) | 0.79 (0.60, 0.98) |
| DWI abnormalities | 0.91 (0.86, 0.96) | 0.88 (0.82, 0.93) | 0.67 (0.60, 0.74) |
| **Numerical Variables** | **Spearman's Correlation [number of entries]** | | |
| Any brain infarct | 0.94 (0.86, 1.0) [79] | 0.95 (0.82, 1.0) [80] | 1.00 (1.00, 1.0) [42] |
| Cortical infarcts | 1.00 (1.0, 1.0) [67] | 0.93 (0.86, 1.0) [71] | 0.96 (0.90, 1.0) [38] |
| Lacunes | 0.93 (0.84, 1.0) [132] | 0.91 (0.79, 1.0) [155] | 0.82 (0.66, 0.98) [55] |
| Cerebellar infarcts | 0.94 (0.85, 1.0) [45] | 0.92 (0.83, 1.0) [51] | 0.86 (0.71, 1.0) [23] |
| Microbleeds | 1.00 (1.0, 1.0) [174] | 1.00 (1.00, 1.0) [180] | 0.97 (0.93, 1.0) [68] |

Table S10: Few-shot information extraction performance using fixed examples, randomly selected examples, structural (Jaccard) similarity-based examples and semantic similarity-based examples measured using F1-Score for categorical and Spearman's correlation for numerical variables.

| Selection method | Zero-shot | Random | Fixed | Structural Similarity | Semantic Similarity |
|---|---|---|---|---|---|
| **Categorical Variables** | **F1-Score** | | | | |
| **Atrophy Measures** | | | | | |
| MTA left | 0.83 (0.66, 1.0) | 0.86 (0.79, 0.92) | 0.86 (0.71, 1.0) | 0.85 (0.69, 1.0) | 0.87 (0.70, 1.0) |
| MTA right | 0.91 (0.80, 1.0) | 0.94 (0.85, 1.04) | 0.96 (0.84, 1.0) | 0.98 (0.96, 0.99) | 0.97 (0.89, 1.0) |
| GCA overall | 0.79 (0.76, 0.82) | 0.80 (0.67, 0.94) | 0.85 (0.72, 0.99) | 0.88 (0.81, 0.95) | 0.86 (0.75, 0.97) |
| GCA frontal | 0.75 (0.64, 0.86) | 0.75 (0.63, 0.87) | 0.86 (0.75, 0.96) | 0.87 (0.83, 0.91) | 0.84 (0.73, 0.95) |
| GCA temporal | 0.71 (0.67, 0.75) | 0.73 (0.60, 0.85) | 0.84 (0.72, 0.95) | 0.83 (0.79, 0.88) | 0.82 (0.75, 0.88) |
| GCA occipital | 0.58 (0.47, 0.68) | 0.62 (0.49, 0.75) | 0.73 (0.54, 0.91) | 0.72 (0.63, 0.80) | 0.71 (0.61, 0.81) |
| GCA parietal | 0.72 (0.67, 0.77) | 0.74 (0.62, 0.85) | 0.80 (0.69, 0.92) | 0.82 (0.72, 0.92) | 0.82 (0.70, 0.94) |
| **Vascular Markers** | | | | | |
| Fazekas score | 0.90 (0.86, 0.93) | 0.90 (0.77, 1.02) | 0.90 (0.81, 1.0) | 0.92 (0.86, 0.97) | 0.91 (0.86, 0.96) |
| Any brain infarct | 0.80 (0.78, 0.82) | 0.81 (0.76, 0.86) | 0.83 (0.74, 0.92) | 0.86 (0.81, 0.91) | 0.84 (0.80, 0.89) |
| Cortical infarcts | 0.71 (0.65, 0.78) | 0.74 (0.68, 0.81) | 0.78 (0.69, 0.86) | 0.79 (0.75, 0.84) | 0.78 (0.73, 0.84) |
| Lacunes | 0.83 (0.80, 0.86) | 0.84 (0.80, 0.88) | 0.85 (0.77, 0.92) | 0.89 (0.85, 0.93) | 0.87 (0.84, 0.91) |
| Cerebellar infarcts | 0.65 (0.61, 0.69) | 0.68 (0.61, 0.76) | 0.74 (0.68, 0.81) | 0.79 (0.73, 0.86) | 0.75 (0.71, 0.78) |
| Splinter infarcts | 0.61 (0.56, 0.65) | 0.73 (0.66, 0.80) | 0.77 (0.72, 0.82) | 0.83 (0.81, 0.86) | 0.80 (0.74, 0.85) |
| Microbleeds | 0.93 (0.91, 0.95) | 0.95 (0.93, 0.96) | 0.95 (0.93, 0.97) | 0.94 (0.93, 0.96) | 0.94 (0.92, 0.96) |
| SWI abnormalities | 0.73 (0.64, 0.81) | 0.72 (0.66, 0.77) | 0.66 (0.45, 0.87) | 0.75 (0.69, 0.80) | 0.71 (0.61, 0.82) |
| Superficial Siderosis | 0.95 (0.87, 1.0) | 0.83 (0.52, 1.15) | 0.93 (0.81, 1.0) | 0.87 (0.58, 1.0) | 0.88 (0.64, 1.0) |
| DWI abnormalities | 0.91 (0.86, 0.96) | 0.90 (0.85, 0.96) | 0.87 (0.60, 1.0) | 0.93 (0.87, 0.98) | 0.93 (0.89, 0.97) |
| **Numerical Variables** | **Spearman's Correlation [number of entries]** | | | | |
| Any brain infarct | 0.94 (0.86, 1.0) [79] | 0.86 (0.71, 1.0) [182] | 0.88 (0.79, 0.97) [203] | 0.80 (0.57, 1.0) [198] | 0.85 (0.59, 1.0) [196] |
| Cortical infarcts | 1.00 (1.0, 1.0) [67] | 0.93 (0.86, 1.0) [181] | 0.94 (0.86, 1.0) [207] | 0.93 (0.85, 1.0) [204] | 0.94 (0.87, 1.0) [202] |
| Lacunes | 0.93 (0.84, 1.0) [132] | 0.88 (0.77, 0.99) [225] | 0.92 (0.80, 1.0) [234] | 0.86 (0.74, 0.97) [237] | 0.85 (0.70, 0.99) [232] |
| Cerebellar infarcts | 0.94 (0.85, 1.0) [45] | 0.92 (0.87, 0.97) [132] | 0.94 (0.88, 1.0) [167] | 0.91 (0.85, 0.97) [180] | 0.90 (0.82, 0.98) [160] |
| Microbleeds | 1.00 (1.0, 1.0) [174] | 1.0 (0.98, 1.0) [249] | 1.0 (0.98, 1.0) [252] | 1.0 (0.98, 1.0) [246] | 1.0 (0.99, 1.0) [248] |

## H. Qualitative examples

Table S11: Illustrative examples with reports, their manual annotations and the output of LLaMA 3.1 obtained using few-shot prompting with structural similarity-based example selection.

| Report (Translated with LLaMA 3.1) | Variable | Manual extraction | Model output |
|---|---|---|---|
| **Example 1** | | | |
| **Medical History:** Acute depression in [date X] with cognitive complaints, followed by gradual recovery. Elsewhere interpreted as possible Alzheimer's.<br>**Question:** Kindly re-evaluation of the MRI brain scan of [date Y], degree of atrophy. **Description:** Re-evaluation of MRI examination performed elsewhere on [date Y], without contrast. Infratentorial normal volume of the brainstem and cerebellar brain parenchyma. Supratentorial thin ventricular system. GCA 0-1 in all brain lobes. MTA 0 bilaterally. Punctiform subcortical white matter lesions, Fazekas 1. No cortical or lacunar infarcts. No diffusion restrictions. No microbleeds.<br>**Conclusion:** No notable atrophy. No pronounced vascular lesions.<br><br>**Original report:**<br>Klinische Gegevens: vrij acute depressie met cognitieve klachten, daarna geleidelijk herstel. Elders geduid als mogelijke Alzheimer. Vraagstelling: Gaarne herbeoordeling MRI-Cerebrum d.d. [date X], mate van atrofie? Beschrijving: Herbeoordeling van MRI-onderzoek verricht elders op [date Y]. Betreft een onderzoek zonder contrast. Infratentorieel normaal volume van de hersenstam en cerebellaire hersenparenchym. Supratentorieel slank ventrikelsysteem. GCA 0-1 in alle hersenkwabben. MTA 0 beiderzijds. Punctiforme subcorticale witte stoflaesies, Fazekas 1. Geen corticale of lacunaire infarcten. Geen diffusie restricties. Geen microbloedingen. Conclusie: Geen opvallende atrofie. Geen uitgesproken vasculaire laesies. | MTA left | 0 | 0 |
| | MTA right | 0 | 0 |
| | GCA overall | 0-1 | 0-1 |
| | GCA frontal | 0-1 | 0-1 |
| | GCA temporal | 0-1 | 0-1 |
| | GCA occipital | 0-1 | 0-1 |
| | GCA parietal | 0-1 | 0-1 |
| | Fazekas score | 1 | 1 |
| | Any brain infarcts | Absent | Absent |
| | Cortical infarcts | Absent | Absent |
| | Lacunes | Absent | Absent |
| | Cerebellar infarcts | Absent | Absent |
| | Splinter infarcts | Absent | Absent |
| | Microbleeds | Absent | Absent |
| | SWI abnormalities | Missing | Missing |
| | Presence of siderosis | No | No |
| | DWI abnormalities | Absent | Absent |
| | # any brain infarct | 0 | 0 |
| | # cortical infarcts | 0 | 0 |
| | # lacunes | 0 | 0 |
| | # cerebellar infarcts | 0 | 0 |
| | # microbleeds | 0 | 0 |
| | Infarct location | Missing | Missing |
| | Microbleed location | Missing | Missing |
| **Example 2** | | | |
| **CLINICAL FINDINGS:** Second opinion due to cognitive complaints. No strong suspicion of dementia syndrome anamnestically.<br>**QUESTION:** Re-evaluation of external imaging MRI-Cerebrum [date X] and PET-CT [date Y]. Atrophy? | MTA left | 0 | 0 |
| | MTA right | 0 | 0 |
| | GCA overall | 0-1 | 0-1 |
| | GCA frontal | **0** | **0-1** |

<table>
<tr><td rowspan="21">**DESCRIPTION**: Re-evaluation of MRI brain performed elsewhere on [date X]. There is a normal volume of infratentorial and supratentorial brain parenchyma, GCA 0-1. Slender ventricular system. There are no white matter lesions. There are no indications of past infarcts. MTA 0 bilaterally. No indications of precuneus atrophy. No diffusion restrictions. No microbleedings.<br>**CONCLUSION:** No pattern of atrophy. No vascular lesions. No structural abnormalities.<br><br>**Original report**:<br>KLINISCHE GEGEVENS: 2nd opinion vanwege cognitieve klachten. Anamnestisch geen sterke verdenking op dementieel syndroom. VRAAGSTELLING: Herbeoordeling externe beeldvorming MRI-Cerebrum [date X] en PET-CT [date Y]. Atrofie? BESCHRIJVING: Herbeoordeling van MRI hersenen verricht elders op [date X]. Er is een normaal volume van infratentorieel en supratentorieel hersenparenchym, GCA 0-1. Slank ventrikelsysteem. Er zijn geen witte stoflaesies. Er zijn geen aanwijzingen voor doorgemaakte infarcten. MTA 0 beiderzijds. Geen aanwijzingen voor precuneus atrofie. Geen diffusie restricties. Geen microbloedingen. CONCLUSIE: Geen patroon van atrofie. Geen vasculaire laesies. Geen structurele afwijkingen</td><td>GCA temporal</td><td>**0**</td><td>**0-1**</td></tr>
<tr><td>GCA occipital</td><td>**0**</td><td>**0-1**</td></tr>
<tr><td>GCA parietal</td><td>**0**</td><td>**0-1**</td></tr>
<tr><td>Fazekas score</td><td>**0**</td><td>**Missing**</td></tr>
<tr><td>Any brain infarcts</td><td>**Missing**</td><td>**Absent**</td></tr>
<tr><td>Cortical infarcts</td><td>**Missing**</td><td>**Absent**</td></tr>
<tr><td>Lacunes</td><td>**Missing**</td><td>**Absent**</td></tr>
<tr><td>Cerebellar infarcts</td><td>**Missing**</td><td>**Absent**</td></tr>
<tr><td>Splinter infarcts</td><td>**Missing**</td><td>**Absent**</td></tr>
<tr><td>Microbleeds</td><td>Absent</td><td>Absent</td></tr>
<tr><td>DWI abnormalities</td><td>Absent</td><td>Absent</td></tr>
<tr><td>Presence of siderosis</td><td>No</td><td>No</td></tr>
<tr><td>SWI abnormalities</td><td>Missing</td><td>Missing</td></tr>
<tr><td># any brain infarcts</td><td>**-1**</td><td>**0**</td></tr>
<tr><td># cortical infarcts</td><td>**-1**</td><td>**0**</td></tr>
<tr><td># lacunes</td><td>**-1**</td><td>**0**</td></tr>
<tr><td># cerebellar infarcts</td><td>**-1**</td><td>**0**</td></tr>
<tr><td># microbleeds</td><td>0</td><td>0</td></tr>
<tr><td>Infarct location</td><td>Missing</td><td>Missing</td></tr>
<tr><td>Microbleed location</td><td>Missing</td><td>Missing</td></tr>
<tr><td></td><td></td><td></td></tr>
<tr><td colspan="4">**Example 3**</td></tr>
<tr><td rowspan="22">**Clinical information:** Suspicion of FTD. Question of whether there was MS in the past. Request for re-evaluation of MRI [date X and Y]. Do the abnormalities match FTD or MS?<br>**Report:** Re-evaluation of MRI according to dementia protocol compared to the [date Y] study. Minimal periventricular white matter abnormalities. No evident white matter abnormalities elsewhere. Left high frontal and several hypointensities on the gradient echo consistent with small hemorrhagic components (image 23 series 5). No areas with diffusion restriction. GCA score: 1, especially the wide fissure of Silvii. No evident asymmetry. MTA score 4 Fazekas score 0. No evident MS lesions. **Conclusion:** The image is more suggestive of Alzheimer's disease due to the significant hippocampus atrophy. MTA 4 No evident signs of FTD or MS.<br><br>**Original report:**<br>Klinische informatie. Verdenking FTD. Zou in voorgeschiedenis MS hebben. Vraagstelling Graag herbeoordeling MRI [date X and Y]. Afwijkingen passen bij FTD wel bij MS? Verslag. Herbeoordeling MRI volgens dementie protocol vergeleken met onderzoek</td><td>MTA left</td><td>4</td><td>4</td></tr>
<tr><td>MTA right</td><td>4</td><td>4</td></tr>
<tr><td>GCA overall</td><td>1</td><td>1</td></tr>
<tr><td>GCA frontal</td><td>**1**</td><td>**Missing**</td></tr>
<tr><td>GCA temporal</td><td>**1**</td><td>**Missing**</td></tr>
<tr><td>GCA occipital</td><td>**1**</td><td>**Missing**</td></tr>
<tr><td>GCA parietal</td><td>**1**</td><td>**Missing**</td></tr>
<tr><td>Fazekas score</td><td>0</td><td>0</td></tr>
<tr><td>Any brain infarcts</td><td>**Missing**</td><td>**Absent**</td></tr>
<tr><td>Cortical infarcts</td><td>**Missing**</td><td>**Absent**</td></tr>
<tr><td>Lacunes</td><td>**Missing**</td><td>**Absent**</td></tr>
<tr><td>Cerebellar infarcts</td><td>**Missing**</td><td>**Absent**</td></tr>
<tr><td>Splinter infarcts</td><td>**Missing**</td><td>**Absent**</td></tr>
<tr><td>Microbleeds</td><td>**Missing**</td><td>**Present**</td></tr>
<tr><td>SWI abnormalities</td><td>Missing</td><td>Missing</td></tr>
<tr><td>Presence of siderosis</td><td>No</td><td>No</td></tr>
<tr><td># any brain infarcts</td><td>**-1**</td><td>**0**</td></tr>
<tr><td># cortical infarcts</td><td>**-1**</td><td>**0**</td></tr>
<tr><td># lacunes</td><td>**-1**</td><td>**0**</td></tr>
<tr><td># cerebellar infarcts</td><td>**-1**</td><td>**0**</td></tr>
<tr><td></td><td></td><td></td></tr>
<tr><td></td><td></td><td></td></tr>
</table>

| | | | |
|---|---|---|---|
| uit [date Z]. Geringe periventriculaire wittestof afwijkingen. Elders geen evidente witte stofafwijkingen. Links hoog frontaal enkele hypo-intensiteiten op de gradient echo passend bij kleine bloedingscomponenten (beeld 23 serie 5). Geen gebieden met diffusierestrictie. GCA score: 1 Met name wat wijde fissura van Silvii. Geen evidente asymmetrie. MTA score 4 Fazekas score 0. Geen evidente MS laesies. Conclusie: Beeld eerder passend bij Alzheimer disease gezien de forse hippocampus atrofie. MTA 4 Geen evidente tekenen van FTD of MS. | # microbleeds | **-1** | **0** |
| | Infarct location | missing | Missing |
| | Microbleed location | **missing** | **left high frontal** |
| **Example 4** | | | |
| **Clinical Data**: [sensitive] **Question**: atrophy, hippocampal atrophy, infarct? RIP? metastases? subdural hematoma? **Description**: MRI brain according to metastasis protocol. Evaluation is complicated by movement artifacts. Cortical atrophy: Left: Frontal GCA: 1 Parietal GCA: 2 Occipital GCA: 1 Temporal GCA: 2-3 Right: Frontal GCA: 1 Parietal GCA: 2 Occipital GCA: 1 Temporal GCA: 2-3 **Medial temporal lobe atrophy (MTA): Not formally assessable due to lack of blank T1 weighted series in the correct plane. Based on the available images, there is moderate mesotemporal atrophy on the right and mild mesotemporal atrophy on the left**. Aspect cerebellum: No evident atrophy. Aspect brainstem: No evident atrophy. Ventricular system: Central liquor system is widened, this is in proportion to the peripheral liquor spaces and the white matter abnormalities. White matter abnormalities: Bilateral extensive periventricular and to a lesser extent subcortical, Fazekas 2. Less than 25% of the white matter is involved. Lacunar or strategic infarcts: None DWI diffusion restriction: None No positive indications for bleeding, however not assessable for microbleeding due to lack of gradient echo or Swan sequence. Quantitative / Quantib: Not possible due to lack of 3D T1 series without contrast. Other: No adjacent lesions, no suspicion of metastases (due to movement artifacts a very small pointiform adjacent sign cannot be ruled out). **Conclusion:** [...]<br><br>**Original report:**<br>Klinische Gegevens: [sensitive] Vraagstelling: atrofie, hippocampusatrofie, infarct? rip? | MTA left | **Missing** | **2** |
| | MTA right | **Missing** | **1** |
| | GCA overall | 2.5 | 2.5 |
| | GCA frontal | 1 | 1 |
| | GCA temporal | 2.5 | 2.5 |
| | GCA occipital | 1 | 1 |
| | GCA parietal | 2 | 2 |
| | Fazekas score | **2** | **2** |
| | Any brain infarcts | Absent | Absent |
| | Cortical infarcts | Absent | Absent |
| | Lacunes | Absent | Absent |
| | Cerebellar infarcts | Absent | Absent |
| | Splinter infarcts | **Absent** | **Missing** |
| | Microbleeds | Missing | Missing |
| | SWI abnormalities | Missing | Missing |
| | Presence of siderosis | No | No |
| | DWI abnormalities | Absent | Absent |
| | # any brain infarcts | **0** | **-1** |
| | # cortical infarcts | **0** | **-1** |
| | # lacunes | **0** | **-1** |
| | # cerebellar infarcts | **0** | **-1** |
| | # microbleeds | -1 | -1 |
| | Infarct location | Missing | Missing |
| | Microbleed location | Missing | Missing |
| metastasen? subduraal hematoom?  Beschrijving: MRI hersenen volgens metastase protocol. Beoordeling wordt bemoeilijkt door bewegingsartefacten.  Corticale atrofie:  Links: Frontaal GCA: 1 Parietaal GCA: 2 Occipitaal GCA: 1 Temporaal GCA: 2-3 Rechts: Frontaal GCA: 1 Parietaal GCA: 2 Occipitaal GCA: 1 Temporaal GCA: 2-3  Medial temporal lobe atrofie (MTA): Niet formeel te beoordelen door ontbreken van blanco T1 gewogen serie in het juiste vlak. Op basis van de beschikbare beelden is er rechts lichte en links een matige mesotemporale atrofie.  Aspect cerebellum: Geen evidente atrofie.  Aspect hersenstam: Geen evidente atrofie. Ventrikelsysteem: Centrale liquorsysteem is verwijd, dit is in verhouding tot de perifere liquorruimten en de witte stof afwijkingen.  Witte stofafwijkingen: Beiderzijds uitgebreid periventriculair en in mindere mate subcorticaal, Fazekas 2. Net minder dan 25% van de witte stof is betrokken.  Lacunaire of strategische infarcten: Geen  DWI diffusierestrictie: Geen  Geen positieve aanwijzingen voor bloedingen, echter niet te beoordelen op microbloedingen bij ontbreken van gradiëntecho of Swan sequentie .  Kwantitatief / Quantib: Niet mogelijk bij ontbreken 3D T1 serie zonder contrast.  Overig: Geen aankleurende laesies, geen verdenking op metastasen (door de bewegingsartefacten kan een zeer kleine puntiforme aankleuring niet worden uitgesloten). Conclusie:  - [...] | | | |

## I. Confusion matrices

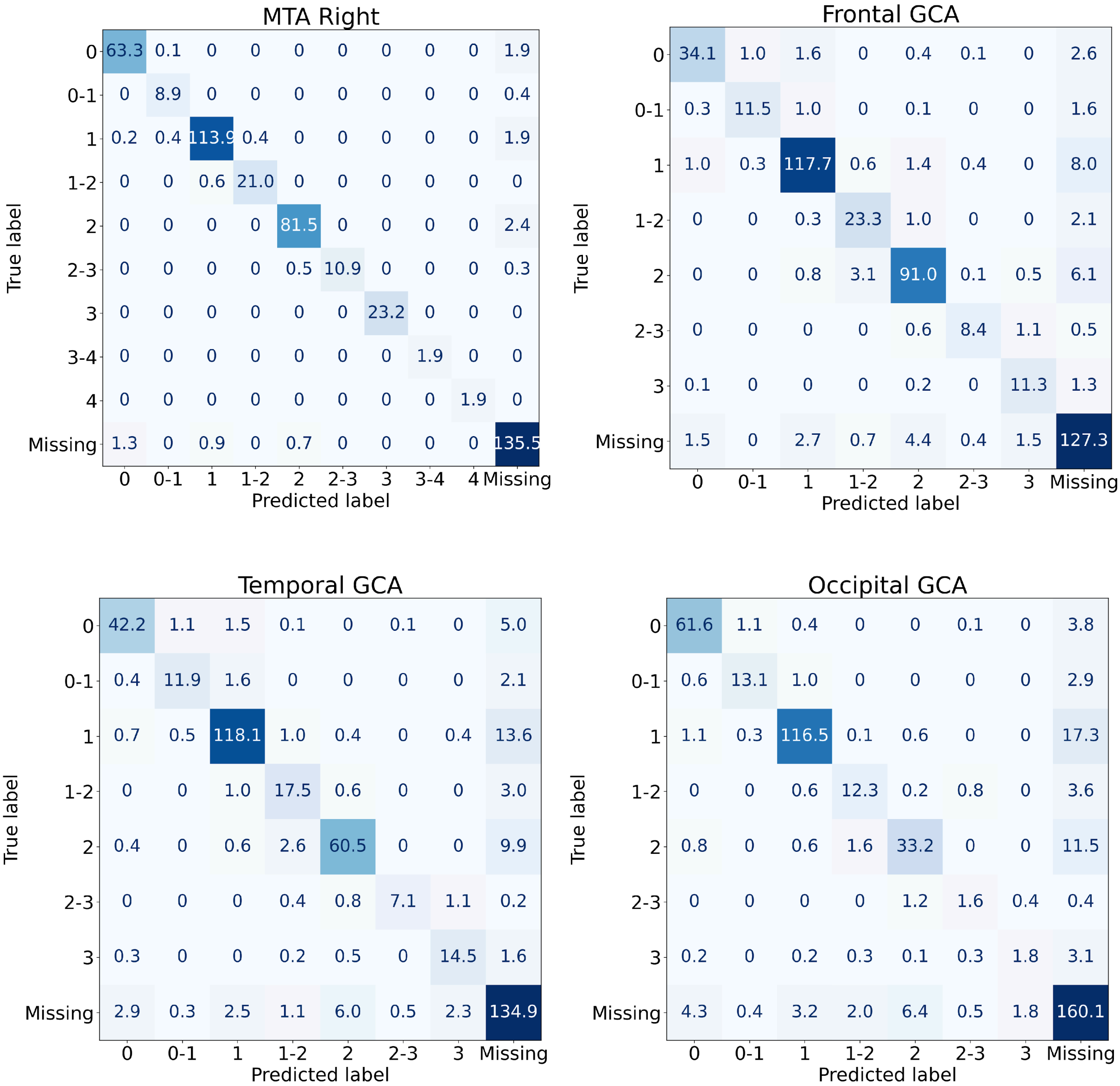
MTA Right
True label
Predicted label
0 0-1 1 1-2 2 2-3 3 3-4 4 Missing
Frontal GCA
True label
Predicted label
0 0-1 1 1-2 2 2-3 3 Missing
Temporal GCA
True label
Predicted label
0 0-1 1 1-2 2 2-3 3 Missing
Occipital GCA
True label
Predicted label
0 0-1 1 1-2 2 2-3 3 Missing

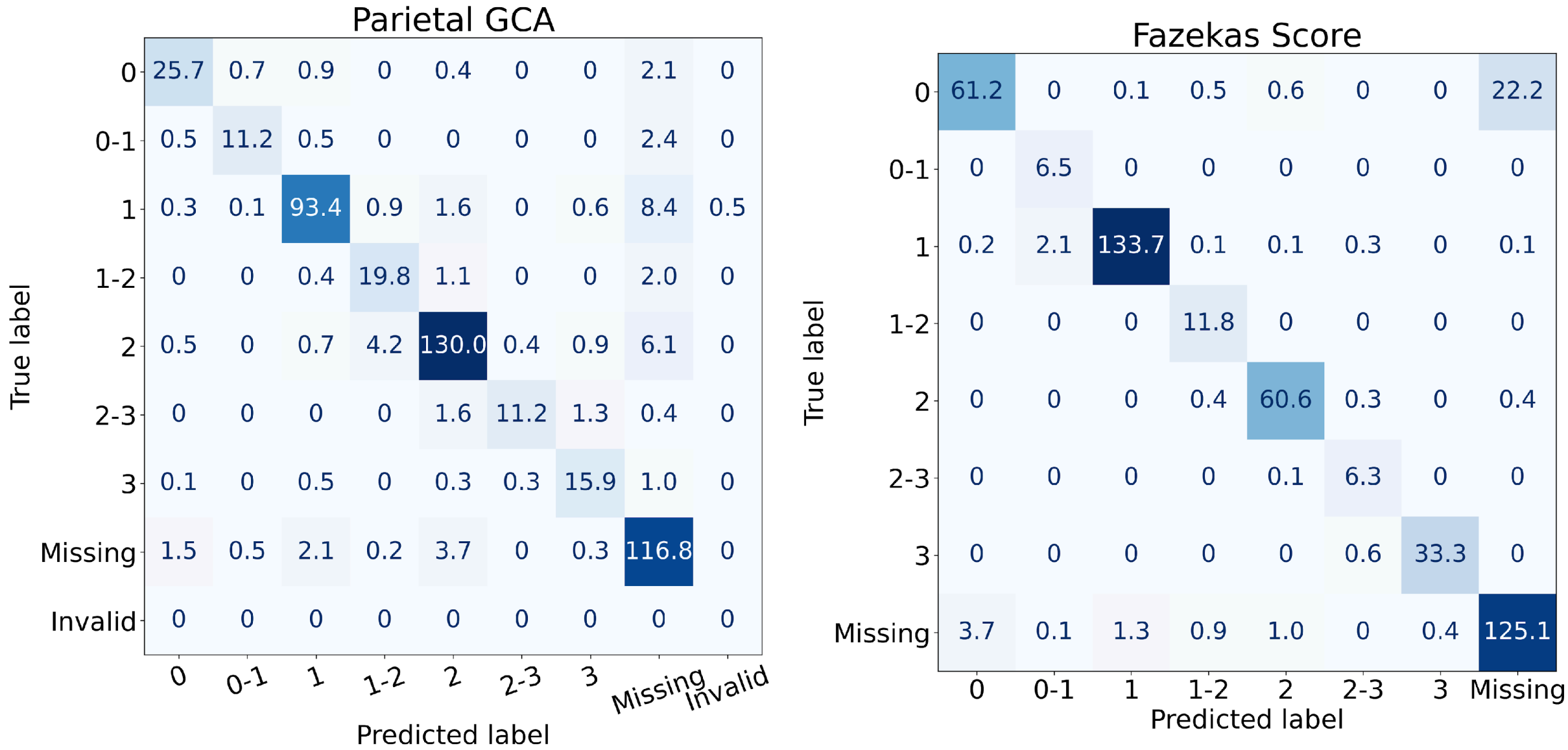

Figure S3: Confusion matrices of atrophy visual rating scores (MTA, GCA, and Fazekas) obtained using few-shot prompting with structural similarity-based example selection.

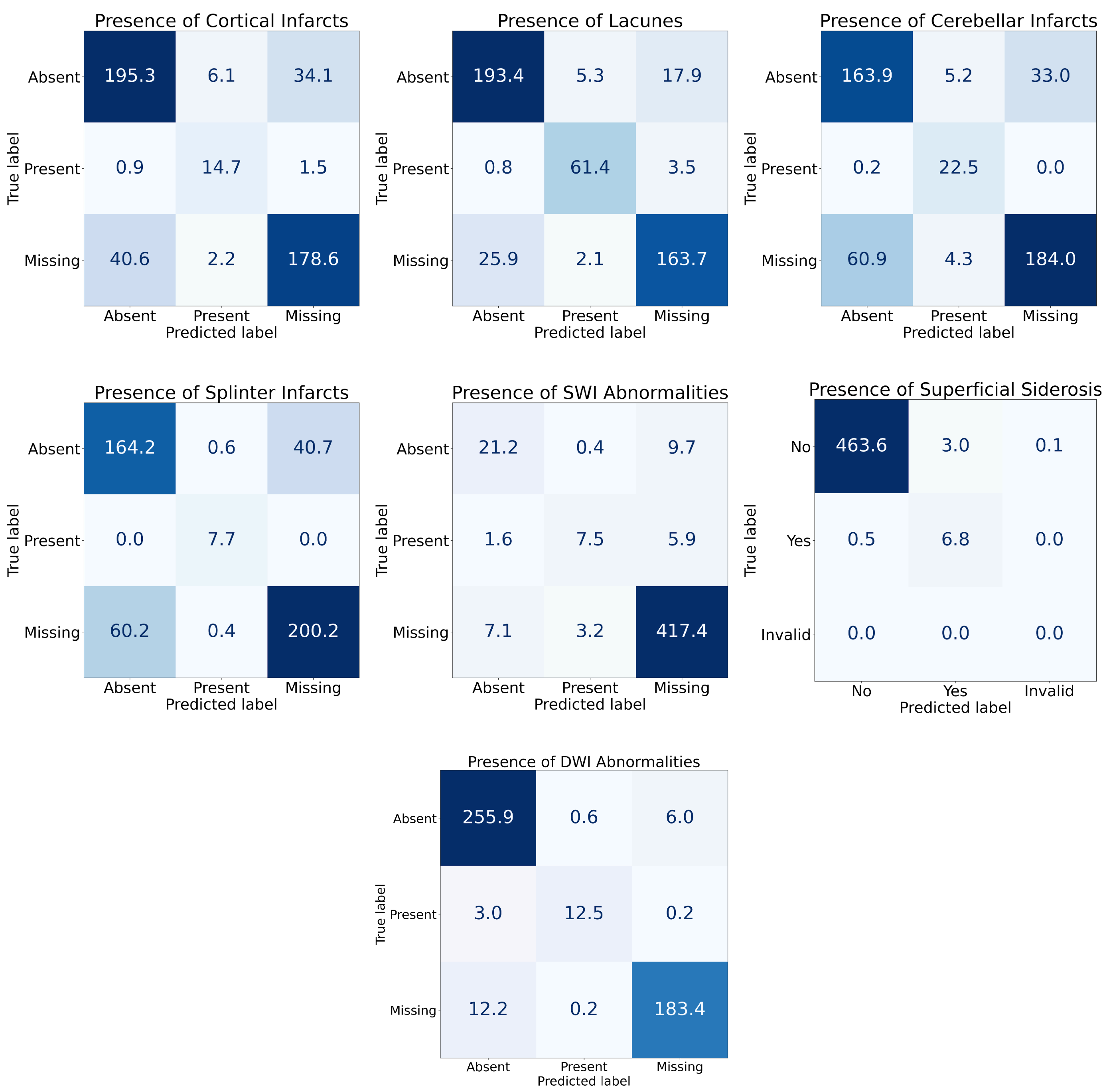

Figure S4: Confusion matrices of atrophy measures obtained using few-shot prompting with structural similarity-based example selection.

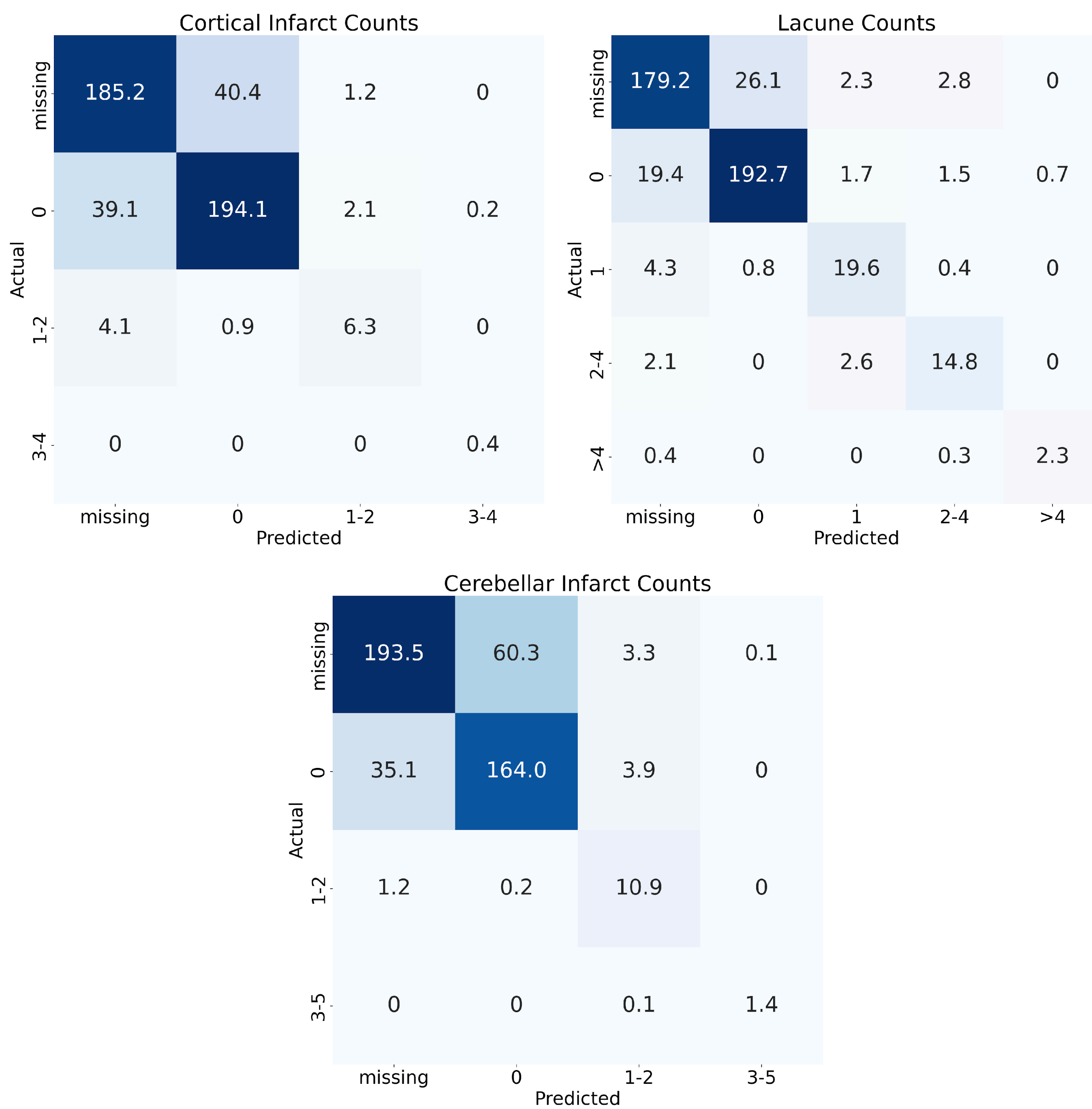

Figure S5: Confusion matrices of lesion counts obtained using few-shot prompting with structural similarity-based example selection.

## J. Radar plots

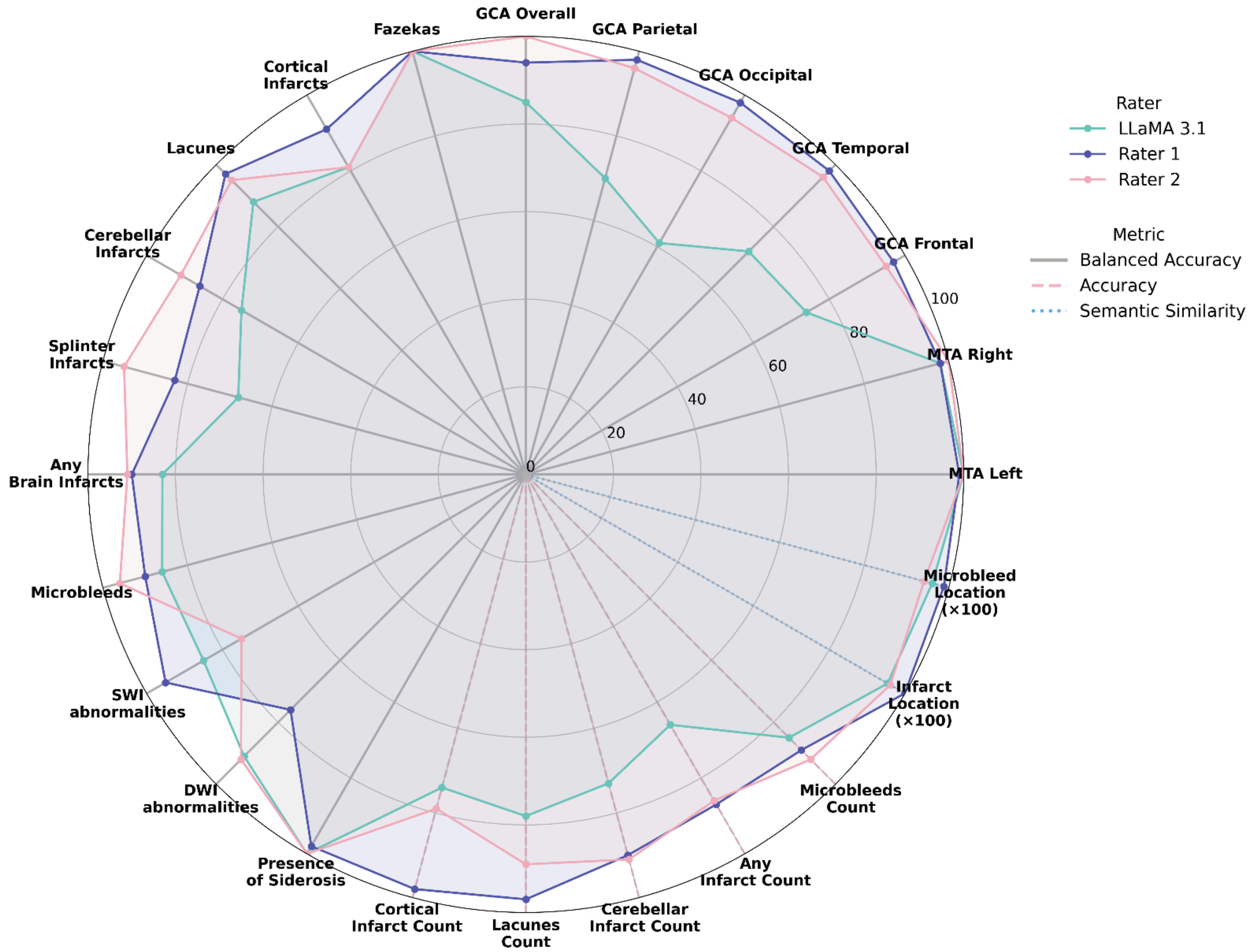

Figure S6: Radar plot of extraction performance per variable achieved by LLaMA 3.1 and each of two human raters as evaluated using the consensus annotations as reference standard. Performance is evaluated on a subset of 100 reports.

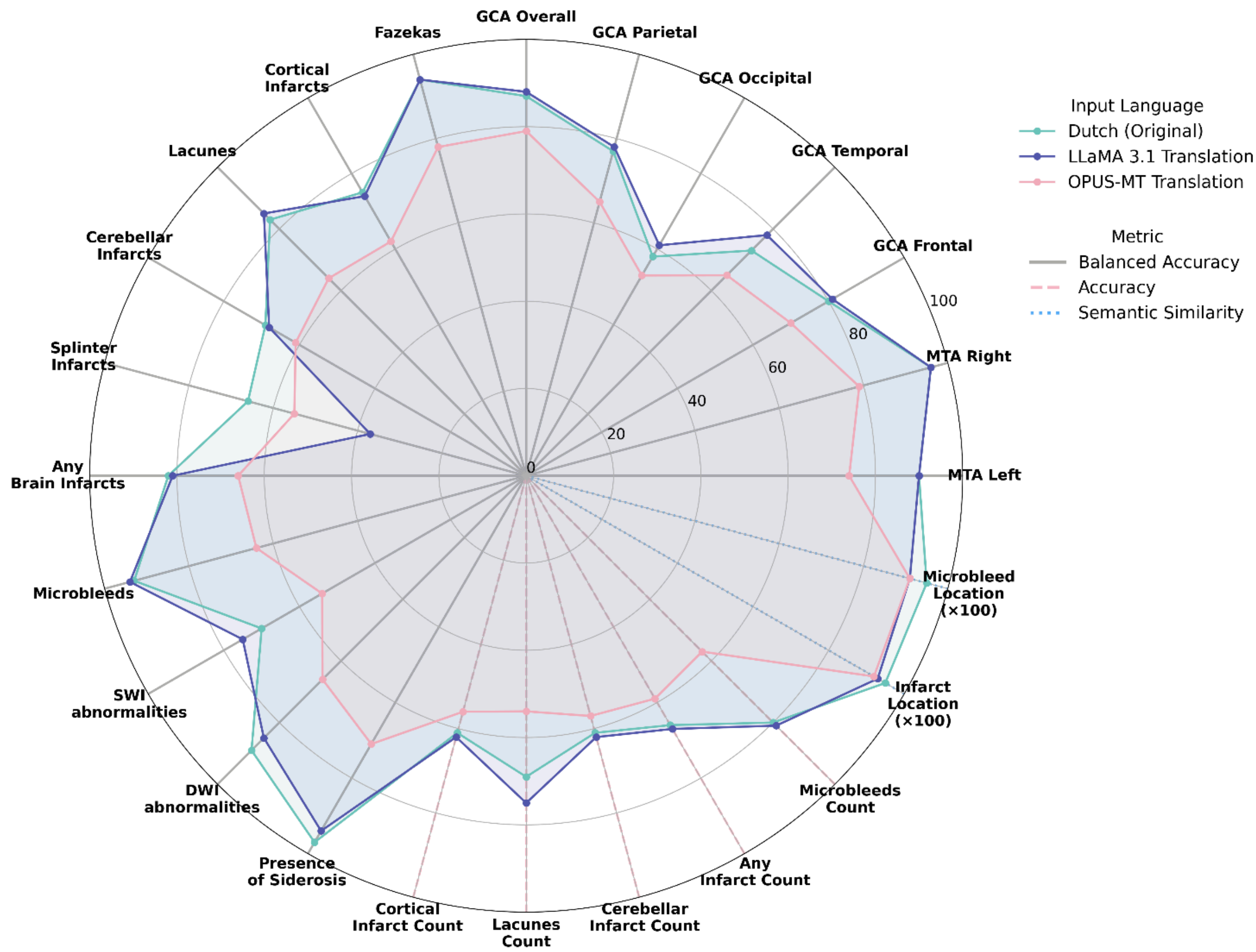

Figure S7: Radar plot of extraction performance per variable achieved using the original Dutch reports, the translation obtained using LLaMA 3.1, and the translation obtained using OPUS-MT as input. Performance is evaluated using the manual annotations as reference standard. Performance is averaged over 10 random splits of 50% of the 947 reports.

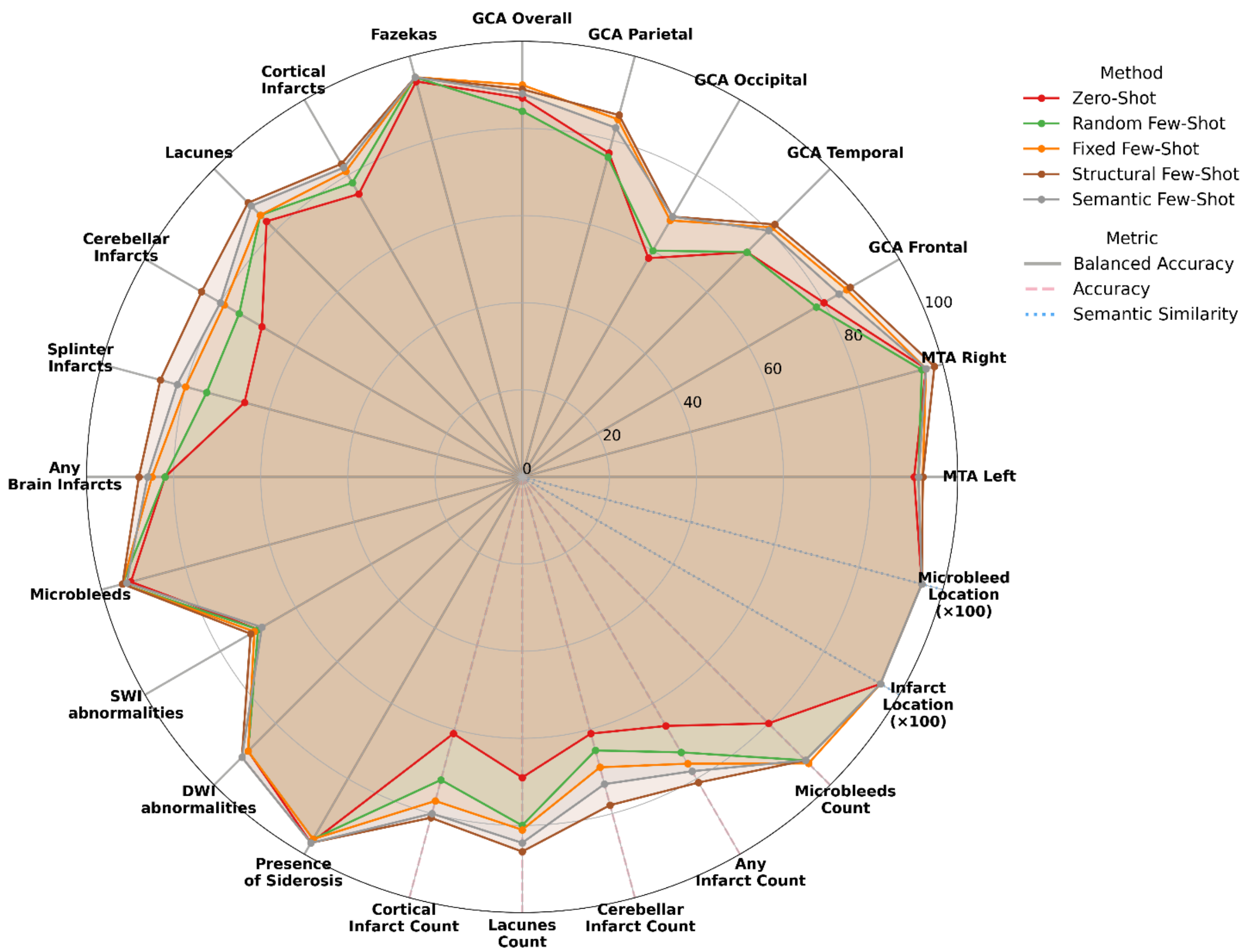

Figure S8: Radar plot of extraction performance per variable achieved using zero-shot prompting and few-shot prompting with random selection, fixed selection, structural similarity-based selection, and semantic semilarity based selection.  Performance is evaluated using the manual annotations as reference standard. Performance is averaged over 10 random splits of 50% of the 947 reports.

## K. Comparison between structural and semantic similarity-based selection

To compare the behavior of the two similarity-based selection methods, we examined the overlap in nearest neighbours selected for each test report. For each of the ten random data splits, we identified the three closest training reports using the structural-similarity measure and the semantic-similarity measure. For every test report, we then quantified how many of the three retrieved neighbours were shared by both methods. This overlap score provides insight into the extent to which structural and semantic similarity identify different few-shot examples. The results are shown in Fig. S7.

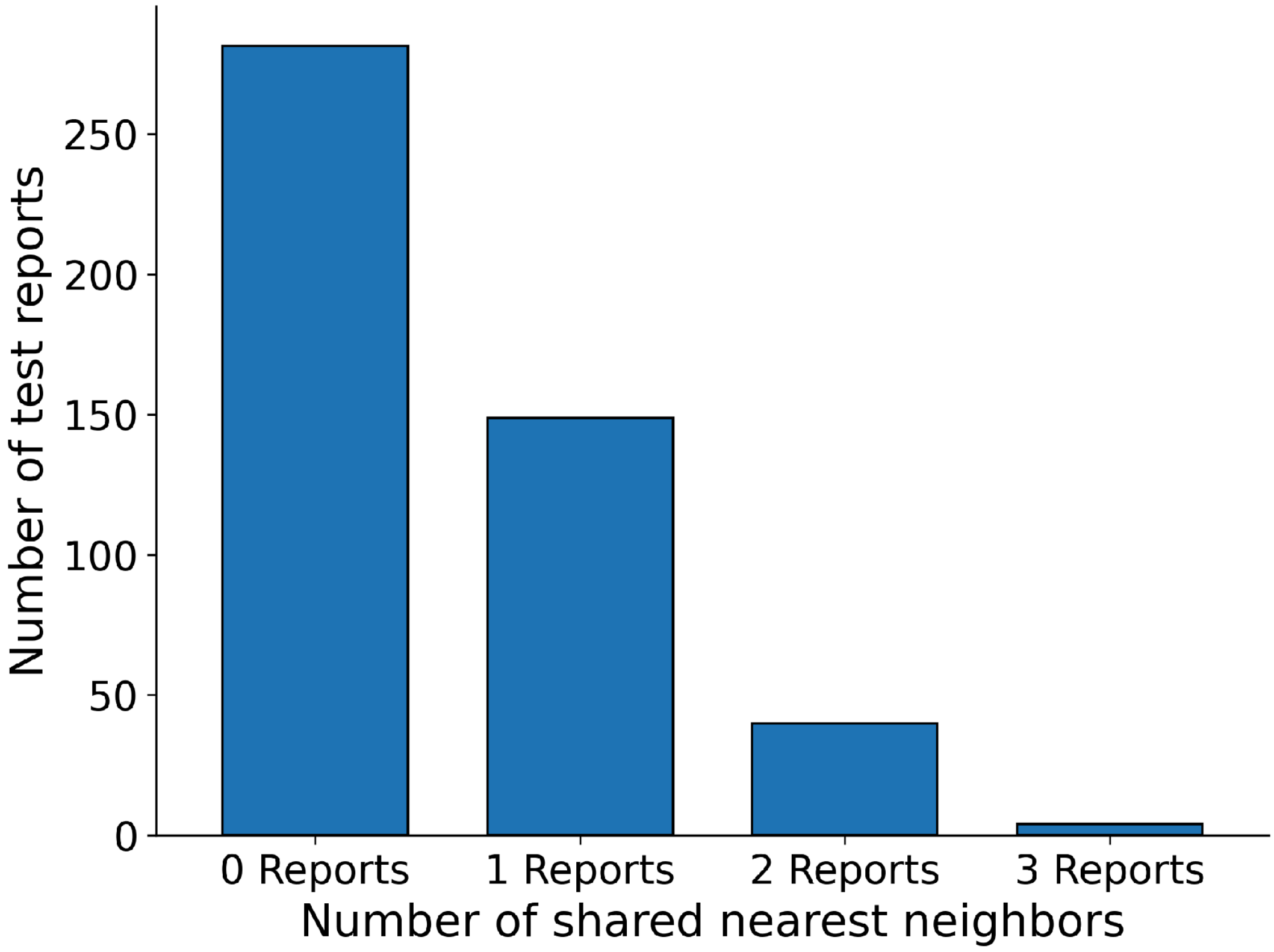

Figure S9: Overlap in nearest-neighbor selection between structural and semantic similarity-based selection. The numbers are averaged over the 10 random splits.

The overlap between structural and semantic similarity is limited. Most test reports share no nearest neighbors across the two methods, indicating that the criteria identify largely different examples in the training set. A substantial minority of reports share one neighbor, while cases with two or three shared neighbors are rare. This pattern suggests that structural and semantic similarity capture distinct aspects of report similarity and therefore retrieve complementary examples rather than converging on the same neighborhood structure.

## L. Sensitivity Analysis: Treatment of the "Missing" Label in Categorical Variables

For categorical variables, the "missing" label (indicating that a finding was not reported) was treated as a distinct class in the primary evaluation, on the grounds that absence of mention is semantically different from an explicitly reported negative finding. To empirically assess the robustness of the reported performance to this choice, we recomputed balanced accuracy under two alternative evaluation schemes:

5. **Excluding missing:** reports with a "missing" reference label were removed, and balanced accuracy was computed on the remaining samples.
6. **Merging missing with negative:** labels coded as "missing" were recoded to the negative finding class ("absent" or “0” depending on the variable, see table S1) in both the reference and predicted labels, and balanced accuracy was computed on the full dataset.

Both schemes were applied for the structured similarity-based few shot prompting method per fold of the 10-fold Monte Carlo cross-validation and averaged across folds, consistent with the primary evaluation protocol. For each variable, we report the mean and standard deviation of balanced accuracy under each scheme, along with the mean per-fold difference (Δ) relative to the original scheme (Table S12).

Table S12: Balanced accuracy (BA) of categorical variables under alternative treatments of the 'missing' label. BA is reported under three evaluation schemes: the original scheme (missing treated as a separate class), excluding reports with a missing reference label, and merging missing into the negative class. Values are mean ± SD across 10 MCCV folds. Δ denotes the mean difference relative to the original scheme.

| Variable | Missing (%) | BA Original (%) | BA Excluding Missing (%) | Δ vs. Original (Excl.) | BA Merging Missing → Negative (%) | Δ vs. Original (Merge) |
|---|---|---|---|---|---|---|
| Atrophy Measures | | | | | | |
| MTA left | 29 | 92 ± 6 | 91 ± 6 | -1 | 91 ± 6 | 0 |
| MTA right | 29 | 98 ± 1 | 98 ± 1 | 0 | 98 ± 1 | 0 |
| GCA overall | 25 | 89 ± 2 | 88 ± 2 | -1 | 88 ± 2 | -1 |
| GCA frontal | 29 | 87 ± 2 | 86 ± 3 | -1 | 87 ± 2 | 0 |
| GCA temporal | 32 | 82 ± 2 | 81 ± 2 | -1 | 82 ± 2 | 0 |
| GCA occipital | 38 | 69 ± 4 | 66 ± 5 | -3 | 67 ± 5 | -3 |
| GCA parietal | 26 | 86 ± 3 | 85 ± 3 | -1 | 86 ± 3 | 0 |
| Vascular Markers | | | | | | |
| Fazekas score | 28 | 95 ± 1 | 95 ± 1 | 0 | 99 ± 1 | +4 |
| Any brain infarct | 43 | 88 ± 2 | 92 ± 1 | +4 | 97 ± 1 | +9 |
| Cortical infarcts | 47 | 83 ± 2 | 85 ± 4 | +1 | 92 ± 4 | +9 |
| Lacunes | 40 | 89 ± 1 | 91 ± 2 | +2 | 96 ± 1 | +6 |
| Cerebellar infarcts | 53 | 85 ± 2 | 90 ± 2 | +5 | 99 ± 1 | +14 |
| Splinter infarcts | 55 | 86 ± 2 | 90 ± 1 | +4 | 100 ± 0 | +14 |
| Microbleeds | 39 | 94 ± 1 | 96 ± 1 | +2 | 99 ± 1 | +5 |
| SWI abnormalities | 90 | 72 ± 4 | 59 ± 5 | -13 | 75 ± 4 | +3 |
| DWI abnormalities | 41 | 91 ± 3 | 89 ± 5 | -2 | 90 ± 5 | 0 |

Superficial siderosis was excluded since it has no “missing” class.

For variables with a low to moderate prevalence of missing labels,  including all atrophy measures (MTA, GCA), Fazekas score, and DWI abnormalities, balanced accuracy under both sensitivity schemes remained within 3% of the primary evaluation, indicating that the reported performance is robust to the handling of missing labels for these variables. For variables with a higher prevalence of missing labels (notably the presence of infarcts with their different subtypes and the presence of microbleeds) merging "missing" with "absent" produced substantially higher balanced accuracy, with increases of up to 14% (cerebellar and splinter infarcts). This confirms that treating "missing" as a separate class represents the more conservative performance estimate, and that merging the two would yield optimistic values by collapsing a genuine model error mode (confusing "missing" with "negative", as characterized in the section "Analysis of common mistakes") into correct predictions by construction. The exclusion scheme produced smaller increases for the same variables, reflecting that removing missing-labeled samples also eliminates a portion of the harder boundary cases. One variable, SWI abnormalities, showed a 13% decrease under the exclude scheme. This reflects the very high prevalence of missing labels for this variable (90%), which leaves a small and heavily imbalanced residual sample after exclusion and produces an unstable estimate; the merge scheme, which retains the full sample, was close to the original (+3%).